\pdfoutput=1

\documentclass{article}

\usepackage{arxiv}

\usepackage[utf8]{inputenc} 
\usepackage[T1]{fontenc}    
\usepackage{hyperref}       
\usepackage{url}            
\usepackage{booktabs}       
\usepackage{amsfonts}       
\usepackage{nicefrac}       
\usepackage{microtype}      
\usepackage{lipsum}		
\usepackage{graphicx}
\usepackage{natbib}
\usepackage{doi}
\usepackage{siunitx}
\usepackage{listings}
\usepackage[usenames]{xcolor}
\usepackage[onehalfspacing]{setspace}
\usepackage{amsmath}

\newcommand{\ltl}{learning to learn}
\newcommand{\LTL}{L2L}
\newcommand{\MS}{\,\si{\ms}}
\newcommand{\HZ}{\,\si{\hertz}}

\definecolor{snsblue}{rgb}{0.12156862745098039, 0.4666666666666667, 0.7058823529411765}
\definecolor{snsgreen}{rgb}{0.17254901960784313, 0.6274509803921569, 0.17254901960784313}

\title{Exploring hyper-parameter spaces of neuroscience models on high performance computers with Learning to Learn}

\author{\href{ https://orcid.org/0000-0001-8869-215X }{\includegraphics[scale=0.06]{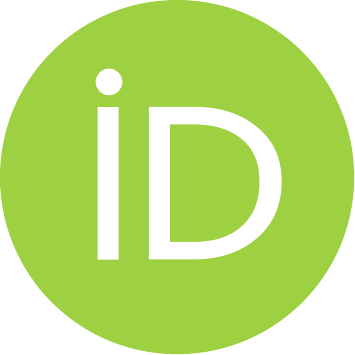}\hspace{1mm}}Alper  Yegenoglu\,$^{1,4, *}$,
  \href{ https://orcid.org/ 0000-0002-7333-9860}{\includegraphics[scale=0.06]{orcid.pdf}\hspace{1mm}} Anand Subramoney\,$^{5}$,
  \href{ https://orcid.org/0000-0002-6249-7169}{\includegraphics[scale=0.06]{orcid.pdf}\hspace{1mm}} Thorsten Hater\,$^{1}$,
  \href{ https://orcid.org/0000-0002-7239-434X}{\includegraphics[scale=0.06]{orcid.pdf}\hspace{1mm}} Cristian Jimenez-Romero\,$^{1}$, \\
  \textbf{
    \href{ https://orcid.org/0000-0002-8996-488X}{\includegraphics[scale=0.06]{orcid.pdf}\hspace{1mm}} Wouter Klijn\,$^{1}$,
    \href{ https://orcid.org/0000-0001-6741-1435}{\includegraphics[scale=0.06]{orcid.pdf}\hspace{1mm}} Aaron Pérez Martín\,$^{1}$,
    \href{ https://orcid.org/0000-0003-0370-4655}{\includegraphics[scale=0.06]{orcid.pdf}\hspace{1mm}} Michiel van der Vlag\,$^{1}$,
    \href{ https://orcid.org/0000-0002-6262-2927}{\includegraphics[scale=0.06]{orcid.pdf}\hspace{1mm}} Michael Herty\,$^{4}$}, \\
  \textbf{
    \href{ https://orcid.org/0000-0001-6933-797X}{\includegraphics[scale=0.06]{orcid.pdf}\hspace{1mm}} Abigail Morrison\,$^{1,2,3}$,
    \href{ https://orcid.org/0000-0002-3168-5394}{\includegraphics[scale=0.06]{orcid.pdf}\hspace{1mm}} Sandra Diaz-Pier\,$^{1}$} \\
  $^{1}$Simulation and Data Lab Neuroscience, Jülich Supercomputing Centre (JSC), \\
  Institute for Advanced Simulation, JARA, \\
  Forschungszentrum Jülich GmbH, Jülich, Germany \\
  $^{2}$Institute of Neuroscience and Medicine (INM-6) \\
  and Institute for Advanced Simulation (IAS-6) \\
  and JARA BRAIN Institute I, \\
  Forschungszentrum Jülich GmbH, Jülich, Germany \\
  $^{3}$Computer Science 3 - Software Engineering,
  \\ RWTH Aachen University, Aachen, Germany\\
  $^{4}$Institute of Geometry and Applied Mathematics, \\
  Department of Mathematics, \\
  RWTH Aachen University, Aachen, Germany\\
  $^{5}$Institute of Neural Computation, \\
  Ruhr University Bochum, Germany \\ [0.3cm]
  Correspondence: \texttt{a.yegenoglu@fz-juelich.de}
}

\date{}

\renewcommand{\headeright}{}
\renewcommand{\undertitle}{}
\renewcommand{\shorttitle}{L2L}

\hypersetup{
pdftitle={Exploring hyper-parameter spaces of neuroscience models on high performance computers with Learning to Learn},
pdfsubject={CS.NE},
pdfauthor={Alper Yegenoglu},
pdfkeywords={simulation, meta learning, hyper-parameter optimization, high performance computing, connectivity generation, parameter exploration},
}

\begin{document}
\maketitle

\begin{abstract}
Neuroscience models commonly have a high number of degrees of freedom and only specific regions within the parameter space are able to produce dynamics of interest. This makes the development of tools and strategies to efficiently find these regions of high importance to advance brain research. Exploring the high dimensional parameter space using numerical simulations has been a frequently used technique in the last years in many areas of computational neuroscience. High performance computing (HPC) can provide today a powerful infrastructure to speed up explorations and increase our general understanding of the model’s behavior in reasonable times.

Learning to learn is a well known concept in machine learning and a specific method for acquiring constraints to improve learning performance.
This concept can be decomposed into a two loop optimization process where the target of optimization can consist of any program such as an artificial neural network, a spiking network, a single cell model or a whole brain simulation.
In this work we present \LTL~as an easy to use and flexible framework to perform hyper-parameter space exploration of neuroscience models on HPC infrastructure.

L2L is an implementation of the \ltl~concept written in Python.  This open-source software allows several instances of an optimization target to be executed with different parameters in a embarrassingly parallel fashion on HPC.
L2L provides a set of built-in optimizer algorithms which makes adaptive and efficient exploration of parameter spaces possible.

Different from other optimization toolboxes, L2L provides maximum flexibility for the way the optimization target can be executed. In this paper we show a variety of examples of neuroscience models being optimized within the L2L framework to execute different types of tasks. The tasks used to illustrate the concept go from reproducing empirical data to learning how to solve a problem in a dynamic environment. We particularly focus on simulations with models ranging from the single cell to the whole brain and using a variety of simulation engines like NEST, Arbor, TVB, OpenAIGym and NetLogo.

\end{abstract}


\section{Introduction}\label{sec:intro}
An essential common tool to most efforts around brain research is the use of algorithms for analysis and simulation.
Specialists have developed a large variety of tools which typically rely on many parameters in order to produce the desired results.
Finding an appropriate configuration of parameters is a highly non-trivial task which usually requires both experience and the patience to comprehensively explore the complex relationships between inputs and outputs.
This problem is common to all input and output formats, as they differ in their type such as images, continuous or discrete signals, experimental data, spiking activity, functional connectivity, etc. In this article we focus on parameter specification for simulation.

In order to address this problem, we present a flexible tool for parameter optimization: L2L.
Initially inspired by the \ltl~concept in the machine learning community, the \LTL~framework is an open-source Python tool\footnote{\url{https://github.com/Meta-optimization/L2L}} which can be used to optimize different workloads.
The flexibility of the framework allows the user to set the target of optimization to be anything which can be executed either from Python or the command line.
The optimization target can also be adaptive and capable of learning, providing a natural way to carry out hyper-parameter optimization.
The L2L framework can be used in local computers as well as on clusters and high performance computing infrastructure.

This manuscript is structured as follows.
First we provide a quick overview on the state of the art for optimization methods and highlight the main differences between those tools and the L2L framework.
In Section~\ref{sec:methods} we provide an overview of the framework’s architecture, its implementation and the way it can be used and extended. We then demonstrate its effectiveness on a variety of use cases focused on neuroscience simulation at different scales (Section~\ref{sec:results}).

\subsection{State of the Art}\label{subsec:state_of_the_art}
In the field of machine learning (ML) the concept of \ltl~(c.f.~Section~\ref{subsec:l2l_concept}) has been well studied.
The learning to learn concept can be decomposed into two components: (a) an inner loop where a program to be optimized, here named the optimizee, executes specific tasks and returns a measure of how well it performs, called the fitness, and (b) an outer loop where an optimizer searches for generalized optimizee parameters (hyper-parameters) that improve the optimizee’s performance over distinct tasks measured by the fitness function. The fitness function is different for each model and tightly linked to the expected transitions in its dynamics.
The optimizee can consist of any program such as an artificial neural network, a spiking network, a single cell model or a whole brain simulation using rate models.
In a recent work~\citet{andrychowicz2016learning} proposed using an LSTM (long short term memory network) with access to the top-level gradients to produce the weight updates for the task LSTM.
The main idea is to replace the gradient descent optimizer of the optimizee with an LSTM as an optimizer.
In this case, the weights of the inner loop network are treated as the hyper-parameters and trained/learned in the outer loop, whilst being kept fixed in the inner loop.
Based on the work of~\cite{andrychowicz2016learning},~\cite{Sachin2017} modified the optimization scheme so that the test error can be incorporated in the optimization step.
Thus, the optimization can be executed in fewer steps which leads to fewer unrollings of the LSTMs and a reduction of the computational burden.
By representing the learning updates of the classifier within the hidden state of the outer-loop optimizer network, the authors acquire a good initialization for the parameters of the inner-loop learner and for further update steps.

For feed-forward networks, Model Agnostic Meta-Learning (MAML) was introduced by~\cite{finn2017model}.
MAML can learn initial parameters for a base-model solving inner-loop-level tasks.
After a few steps of optimization with gradient descent the base-model can generalize well on the validation set, which is unseen but related data from the same class as the training set.
The method can be applied to a vast set of learning problems since the learning itself is agnostic to the inner-loop model.
\citet{finn_MetaLearning_2017} showed that learning the initialisation combined with gradient updates was as powerful as doing learning to learn using a recurrent network.
Several extensions have been proposed to enhance the performance of the learning and computation time~\citep{finn2018probabilistic,finn2019online}.
For example,~\cite{li2017meta} introduce META-SGD, a stochastic gradient optimization method which not only learns the parameter initialization but also the gradient update of the base-model optimization.
However,~\cite{antoniou2018train} lists several issues found with MAML, such as training instabilities due to repeated application of backpropagation through the same network multiple times which leads to gradient issues.
This leads to performance drop in learning and computational overhead.
A gradient-free version of MAML was proposed in~\cite{song2019maml} using evolution strategies to replace the second-order backpropagation used in MAML.
A framework which is model agnostic but doesn't depend on calculating gradients or backpropagating through networks and is not limited to a single optimization algorithm would be highly desirable, specially to address the needs of highly interdisciplinary fields such as neuroscience.

~\cite{cao2019swarm} utilize particle swarm optimization~\citep{kennedy1995particle} to train a meta-optimizer that learns both point-based and population-based optimization algorithms in a continuous manner.
The authors apply a set of LSTMs to train and learn the update formula for a population of samples.
Their learning is based on two attention mechanisms, the feature-level (``intra-particle'') and sample-level (``inter-particle'') attentions.
The intra-particle module reweights every feature based on the hidden state of the corresponding $i$-th LSTM, whereas the inter-particle attention module learns in the update step of the actual particle information from the previous already updated particles.

In a similar manner~\cite{jaderberg2017population} use a parallel population based approach and random search to optimize the hyper-parameters of neural networks.
They randomly sample the initialization of the network parameters and hyper-parameters and every training run is evaluated asynchronously.
If a network is underperforming it is replaced by a more successful network.
Furthermore, by perturbing the hyper-parameters of the replacing network the search space is expanded.
Neural architecture search~\citep{zoph_Neural_2016} and related methods have been shown to be very useful in choosing network architectures for various tasks.
Random search was shown to be surprisingly effective for hyper-parameter searches for a wide variety of tasks~\citep{bergstra_Random_2012}.
Many of the automated hyper-parameter search also fall under the category of Automated Machine Learning or AutoML~\citep{hutter2019automated,he2021automl}.

In the area of computational neuroscience BluePyOpt~\citep{bluepyopt} has represented a robust solution to address optimization problems. Even if it was originally meant to support the optimization of single cell dynamics, BluePyOpt is able to optimize also models at other scales. It makes use of DEAP~\citep{DEAP_JMLR2012} for the optimization algorithms and of SCOOP~\citep{SCOOP_XSEDE2014} to provide parallelization. The target of optimization in BluePyOpt is also quite flexible, it can be any simulator that can be called from Python. This framework can also be used in different infrastructure, from laptops to clusters.
However, the framework only allows execution of optimization targets written in Python.

Deep Learning compatible spiking network library such as NengoDL~\citep{Rasmussen2018} or Norse~\citep{norse2021} are getting more popular.
They are based on modern tensor libraries and can be executed on GPUs which can speed up the simulations.
Although these libraries do not focus on meta-learning they are interesting for solving machine learning tasks using spiking neural networks (SNN).
They can be used to quickly learn the tasks while the hyper-parameters of the SNNs can be optimized in a outer loop.

The L2L framework offers a flexible way to optimize and explore hyper-parameter spaces.
Due to its interface the optimization targets are not restricted to executables with a Python interface offering the possibility to optimize models written in different programming languages.
In our work we focus on neuroscientific use cases, the framework, however, is available for a variety of simulations in different scientific domains.
Furthermore, the framework is agnostic to the inner loop models and thus allows for different types of optimization techniques in the outer loop.
Most of the optimizers adapt population based computational algorithms, which enable parallel executions of optimizees (see Section~\ref{sec:results}).
This helps to optimize for a vast range of parameter ranges.
The error or rather fitness of the inner loop on the absolved tasks is included in the optimization step to update the parameters.
Optimizers such as the Genetic Algorithm or Ensemble Kalman Filter use the fitness in order to rank the individuals and replace underperfoming individuals with more successful ones(e.g.~see Section~\ref{subsec:uc1_nest})).

\section{Methods}\label{sec:methods}
\subsection{Concept of Learning to Learn (L2L)}\label{subsec:l2l_concept}

\begin{figure}[ht]
  \centering
  \includegraphics[width=0.7\textwidth]{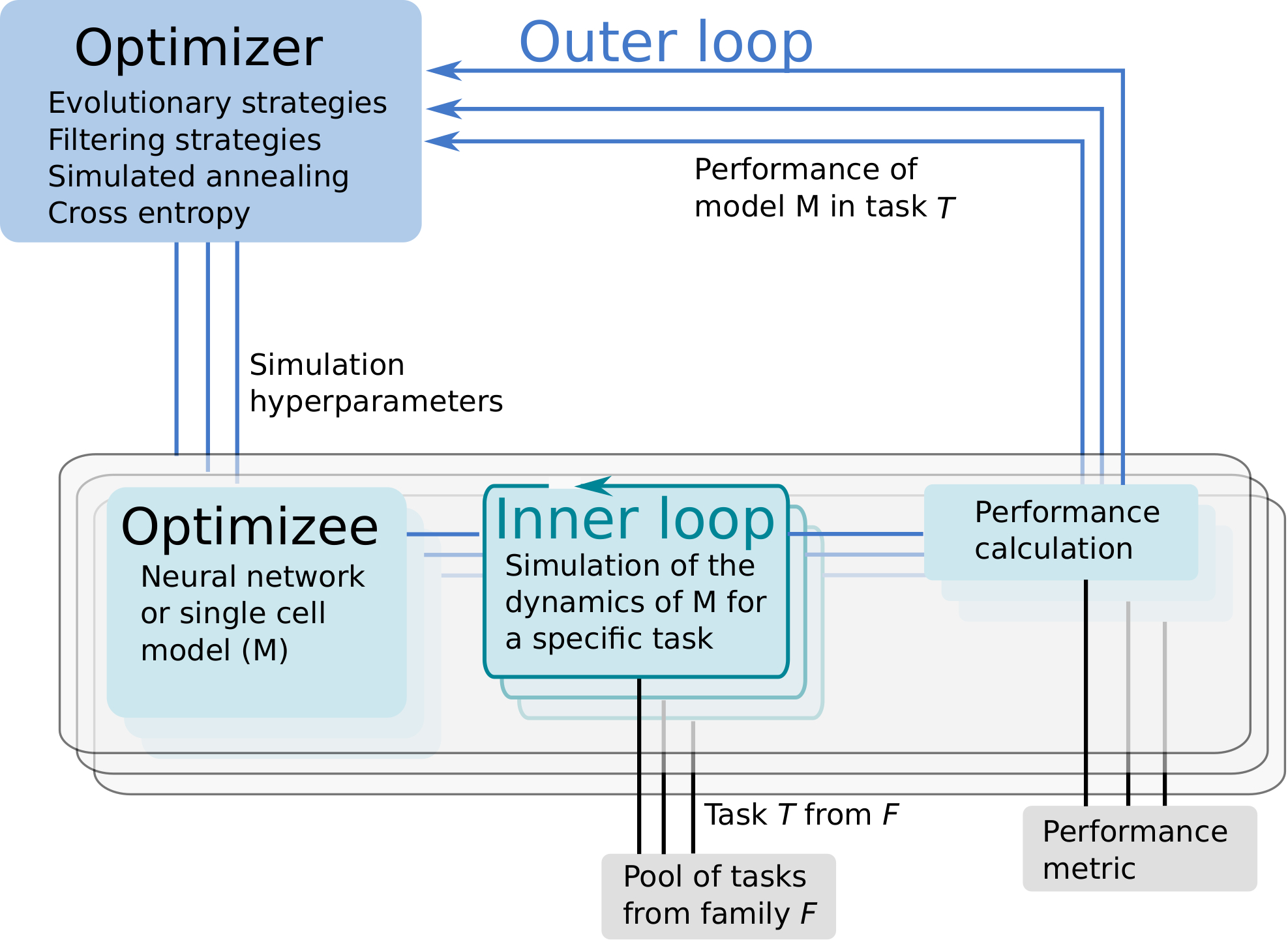}
  \caption{\label{fig:l2l} L2L consists of two loops. In the inner loop, the optimizee, an algorithm with learning capabilities is trained on a family of tasks. A fitness function evaluates the performance of the algorithm. The (hyper-) parameters and the fitness value of the algorithm are send to the optimizer in the outer loop. Several optimization methods are available to optimize the parameters, which are fed back to the optimizee and the algorithm.}
\end{figure}

Learning to learn or meta-learning is a technique to induce learning from experience~\citep{thrun2012learning}.
The \ltl~process consists of two loops, the inner and outer loop (Figure~\ref{fig:l2l}).
In the inner loop, an algorithm with learning capabilities (e.g.~an artificial or spiking neural network, a single cell model or a whole brain simulation using rate models) is executed on a specific task $T$ from a family $\mathcal{F}$ of tasks.

Tasks can range from classification \citep[e.g. MNIST;][]{lecun2010mnist}, see Section~\ref{subsec:uc1_nest}, to identifying parameter regimes that result in specific network dynamics (Sections~\ref{subsec:uc2_arbor},~\ref{subsec:uc4_tvb}) or training agents to autonomously solve optimization problems (Sections~\ref{subsec:uc3_netlogo},~\ref{subsec:uc5_gym}).

The performance of the algorithm over tasks is evaluated with a specifically designed fitness function, which produces a fitness value $f$ or a fitness vector $\mathbf{f}$.
The function is, in general, different for every model but closely connected to the task itself.
Parameters and hyper-parameters, together with the fitness value of the optimizee are sent to the outer loop.
Different optimization techniques, such as evolutionary algorithms, filtering methods or gradient descent can be utilized to optimize the hyper-parameters in order to improve the optimizee's performance.
Afterwards, the hyper-parameters are fed back into the algorithm and a new iteration (i.e.~a new generation) is invoked.
It is important to note that from a technical point of view, the optimizee acts as an orchestrator of the inner loop.
Each optimizee executes a simulation.
Borrowing the terminology from evolutionary algorithms, the parameter set which is optimized is called an individual.
The optimizee accepts (hyper-)parameters from the outer loop and starts the inner loop process to execute the simulation.
Lastly, it calculates the fitness and transmits everything to the optimizer.

\subsection{Parallel Executions in the L2L framework}\label{subsec:l2l_parallel_exec}
In L2L the optimizers apply population based methods which enables simulations to be run in an embarrassingly parallel fashion.
Each individual is initialized independently .
They can be easily distributed on several computing nodes and thus can exploit high-performance computing systems.
The L2L framework supports MPI over several nodes and multi-threading per node.
The number of nodes and cores can be set in the beginning of the run and the L2L framework will automatically take care over the distribution and collection of results.
The next Section~\ref{subsec:l2l_workflow} explains in detail how to set up a simulation run in L2L.

\subsection{Workflow description}\label{subsec:l2l_workflow}
\begin{lstlisting}[language=python, label={code:l2l-template}, caption={Template script to start a L2L run. The optimizee, optimizer are defined. The experiment class is managing the run.},captionpos=b]
from l2l.utils.experiment import Experiment
from l2l.optimizees.optimizee import Optimizee, OptimizeeParameters
from l2l.optimizers.optimizer import Optimizer, OptimizerParameters

experiment = Experiment(root_dir_path='/home/user/L2L/results')
jube_params = {"exec": "srun -n 1 -c 8 --exclusive python"}
traj, all_jube_params = experiment.prepare_experiment(name='L2L-Run',
                                                      log_stdout=True,
                                                      jube_parameter=jube_params)

## Inner loop simulator
# Optimizee class
optimizee = Optimizee(traj)
optimizee_parameters = OptimizeeParameters()

## Outer loop optimizer initialization
optimizer_parameters = OptimizerParameters()
optimizer = Optimizer(traj,
                      optimizee_prepare=optimizee.create_individual,
                      fitness_weights=(1.0,),
                      optimizee_bounding_func=optimizee.bounding_func,
                      parameters=optimizer_parameters)

experiment.run_experiment(optimizee=optimizee,
                          optimizee_parameters=optimizee_parameters,
                          optimizer=optimizer,
                          optimizer_parameters=optimizer_parameters)
experiment.end_experiment(optimizer)
\end{lstlisting}

In L2L the user has to work on two main files.
The first file is the \textbf{run script} which invokes the whole L2L two loop run.
The second file is the \textbf{optimizee}, which operates the simulation in the inner loop.

In the run script the user configures hardware related settings, e.g. if the run is executed on a local computer or on an HPC.
Furthermore, the optimizee and optimizer as well as their parameter options have to be set.
An example code template to start the whole L2L run is shown in Listing~\ref{code:l2l-template}.
Lines 1-3 import the necessary modules, i.e. the experiment, optimizee and the optimizer.
Of course, in a real run the names of the modules and classes have to be adapted to their respective class names, for simplicity we call them here Optimizee and Optimizer.
The \textbf{experiment} class manages the run.
In line 5 the results path is set in the constructor of the class.
The experiment method \texttt{prepare\_experiment} in line 7 prepares the run.
It accepts the name of the run, whether logging should be enabled and the Juelich Benchmarking Environment \citep[JUBE;][]{Speck:901885} parameters.
In L2L, JUBE's functionality was stripped down to submit and manage parallel jobs on HPCs and interacts with the job management system SLURM~\citep{yoo2003slurm}.
The execution directives for the HPC jobs can be seen in line 6.
Here, \texttt{exec} is the indicator command to invoke a run on a supercomputer, followed by a \texttt{srun} directive for SLURM.
In the example, one task (\texttt{-n 1}) should be run on 8 cores (\texttt{-c 8}).
Optimizees and optimizers run as Python executables, which is why the \texttt{python} command is needed here.
If a local run is desired, just the Python command is sufficient, i.e. \lstinline!"exec":"python"! 
Internally, JUBE creates a job script and passes it to SLURM, which then executes the parallel optimizees and the optimizer.
JUBE accepts many more commands for SLURM, but elaborating on all options would go beyond the scope of this work; see the SLURM documentation~\footnote{\url{https://slurm.schedmd.com/}} for a list of executives.
The run script can be executed either as a batch script or as an interactive job on an HPC.

The \textbf{optimizee} is defined in line 13, and requires only the trajectory \texttt{traj}.
The trajectory, modelled after PyPet's trajectory\footnote{\url{https://github.com/SmokinCaterpillar/pypet}}, is a class which holds the history of the parameter space exploration and the results from each execution as well as the parameters to be explored.
\texttt{OptimizeeParameters} is a Python \texttt{namedtuple} object, which accepts the parameters of the optimizee.
For the optimizee, the \texttt{namedtuple} appears as a parameter object and can be accessed as a class variable, i.e. as \texttt{parameters.name}.
The optimizee has access to the trajectory and the parameters object.

In the optimizee three main functions have to be implemented.
\begin{enumerate}
    \item The function \texttt{create\_individual()} defines the individual. Here, the parameters which are going to be optimized need to be initialized and returned as a Python dictionary.
    \item \texttt{simulate()} is the main method to invoke the simulation. The L2L framework is quite flexible about the simulation in the inner loop. It is agnostic with regards to the application carrying out the simulation and only requires that a fitness value or fitness vector is returned.
    \item \texttt{bounding\_func()} is a function which clips parameters before and after the optimization to defined ranges. For example in a spiking neural network it is necessary that delays are strictly positive and greater than zero. The function is applied only on parameters which are defined in \texttt{create\_individual()}.
\end{enumerate}

In a similar fashion, the \textbf{optimizer} is created in line 18.
It requires the optimizer parameters (line 17) and the method \texttt{optimizee.create\_individual},
and if available the bounding function \texttt{optimizee.bounding\_func}.
Additionally, a tuple of weights ( \texttt{fitness\_weights}, here $(1.0,)$) can be given, which weights the optimizee fitness by multiplying those values with the fitness itself.
For example, in case of a two dimensional fitness vector, a tuple of $(1.0, 0.5)$ would weight the first fitness fully and the second one only half.
Most of the  optimizers in the L2L framework perform fitness maximization, but if minimization is required then it suffices to flip the sign of the fitness function that would be used for maximization.
Several optimization techniques are available in the framework, such as cross-entropy, genetic algorithm (GA), evolutionary strategies~\citep{salimans2017evolution}, gradient descent, grid-search, ensemble Kalman filtering~\citep{iglesias2013ensemble} (EnKF), natural evolution strategies~\citep{wierstra2014natural}, parallel tempering and simulated annealing.
The results of the optimizations are automatically saved in a user specified results folder as Python binary files, however users can store result files from within the optimizee in any format they wish.

The method \texttt{run\_experiment} (line 24) requires that the optimizee and the optimizer and their parameters have been defined.
Finally, the \texttt{end\_experiment} method is needed to end the simulation and to stop any logging processes.

\section{Results}\label{sec:results}
In this section we present the results of using L2L to optimize the parameters for a variety of simulation use cases. Every task is executed with a different set of simulation tools, and the interfaces with the simulators also differ between use cases.
We present here 5 use cases. Please see the supplementary material for an additional use case.

\subsection{Use case 1: Digit classification with NEST }\label{subsec:uc1_nest}
The first use case describes digit classification with a spiking neural network (SNN) implemented in the NEST simulator~\citep{gewaltig2007nest}.
The SNN is designed as a reservoir, i.e.~a liquid state machine (LSM, ~\citet{maass2002real}).
The network consists of an input encoding layer, a recurrent reservoir and an output layer as shown in Fig.~\ref{fig:reservoir}. The weights between the reservoir and the output layer are optimised to maximize the classification accuracy.

\subsubsection{Description of the simulation tool}\label{subsubsec:uc1_description}
NEST is a simulator for spiking neural network models. Its primary design focus is the efficient and accurate simulation of point neuron models, in which the morphology of a neuron is abstracted into a single iso-potential compartment; axons and dendrites have no physical extent.
Since NEST supports parallelization with MPI and multi-threading and exhibits excellent scalability, simulations can either be executed on local machines or efficiently be scaled up to large scale runs on HPCs~\citep{jordan2018extremely}.
Our experiments were conducted on the HDF-ML cluster of the Jülich Supercomputing Center using NEST 3.1~\citep{deepu_rajalekshmi_2021_5508805}.

\subsubsection{Optimizee: Spiking reservoir model}\label{subsubsec:uc1_inner}
\begin{figure}[ht]
    \centering
    \includegraphics[width=0.8\linewidth]{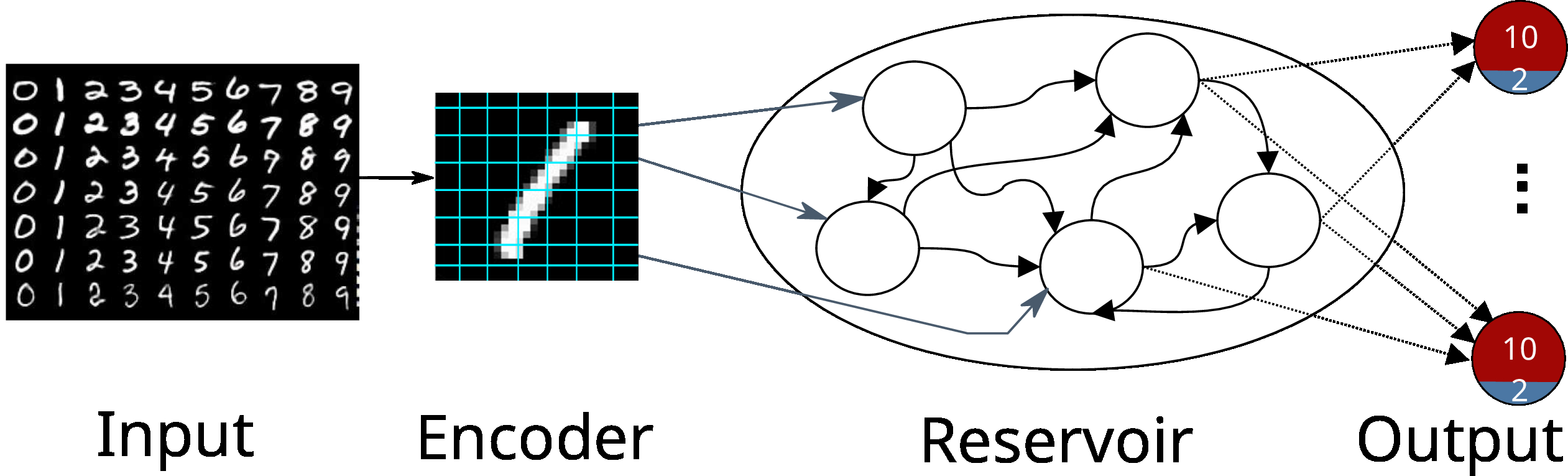}
    \caption{A schematic view of a reservoir network classifying the MNIST dataset. The input image is encoded into firing rates and fed afterwards into the reservoir. The output consists of $10$ excitatory neuron depicted in red and $2$ inhibitory neurons depicted in blue. The highest activity at the output indicates the presented digit.}
    \label{fig:reservoir}
\end{figure}

The network consists of three populations of leaky integrate-and-fire (LIF) neurons, the encoder, the reservoir and the output; see Fig.~\ref{fig:reservoir}.
The input to the network is the set of MNIST digits, encoded into firing rates; the firing rates are proportional to the intensity of the pixels from $0$ to $255$ mapped between $[1, 100]$\HZ.\@
A total of $768$ excitatory neurons receive input from a pixel of the image in a one-to-one connection.
The reservoir has $1600$ excitatory and $400$ inhibitory neurons, while the output has a population of $12$ neurons ($10$ excitatory (red), $2$ inhibitory (blue)) per digit.
The connections in the reservoir are randomly connected, but limited to a maximal outdegree of $6\% \text{ and } 8\%$ for each excitatory and inhibitory neuron.
Each output cluster receives a maximal indegree of $40\%$ of the connections from the reservoir.
The neurons within an output are recurrently connected, while the output clusters don't have connections to each other.
If an input is not presented the network exhibits low spiking activity in all three parts.
The whole network is constructed in the \texttt{create\_individual} function.
The connection weights are sampled from a normal distribution with $\mu=70 \text{ and } \sigma=50$ for the excitatory neurons and $\mu=-90 \text{ and } \sigma=50$ for the inhibitory neurons.

In the simulation (\texttt{simulate} function) a small batch of $10$ different numbers from the same digit is presented to the network for \SI{500}{\ms} per image as spike trains.
Additionally, poissonian noise is added to the network to always maintain a low activity within the reservoir.

Before any image is presented there is a warming up simulation phase lasting for \SI{100}{ms} in order to decay all neuron parameters to their resting values.
Likewise, between every image there is a cooling period of \SI{200}{ms} where no input is shown.
After the simulation is run, the output with the highest spike activity indicates the number of the presented digit.

\subsubsection{Fitness metric}\label{subsubsec:uc1_metric}
In the output we acquire the firing rates of all clusters and apply the softmax function
$$ \sigma(\mathbf{x})_j = \frac{e^{x_j}}{\sum_{k} e^{x_k}}\, , $$
where $\sigma: \mathbb{R}^{k} \rightarrow [0,1]^{k}$ and $\mathbf{x} = (x_{0}, x_{1}, \ldots x_{k}) \in \mathbb{R}^{k}$, $j = 1, \ldots, k$ is the vector of firing rates.
We take the highest value, which indicates the digit the network classified.
Since every image in the dataset has a label we can calculate the loss by applying the mean squared error function to the corresponding label:
\begin{equation}
     \mathcal{L} = \frac{1}{n}\sum^{n}_{i=1}(y_i - \hat{y}_i)^2 \, ,
\end{equation}
with $y_i$ the label and $\hat{y}_i$ the predicted output, encoded as one-hot vectors with a non-zero entry corresponding to the position of the label.
As the optimizer used in the outer loop for this use case is the ensemble Kalman filter, which minimizes the distance between the model output and the training label,
we define the fitness function as $f = 1-\mathcal{L}$ and use it in order to rank individuals (see next Section~\ref{subsubsec:uc1_outer}).
After each presentation of a digit, the fitness and the softmax model output is sent to the optimizer.

\subsubsection{Optimizer: Ensemble Kalman filter}\label{subsubsec:uc1_outer}
The ensemble Kalman filter~\citep[EnKF;][]{iglesias2013ensemble} is the optimization technique we use to update the weights between the reservoir and the output, as described in~\citet{yegenoglu2020ensemble}. Before the optimization, they are normalized to be in the range of $[0,1]$.
The weights from the reservoir to the output are concatenated to construct one individual.
In total, $98$ individuals go into the optimization.
Each individual has $28800$ weights.
To specify in terms of the EnKF setting, the set of ensembles are the network weights, the observations are the softmax model outputs.
In~\citep{yegenoglu2020ensemble} it was shown that around $100$ ensembles are required to reach at least chance level on the MNIST dataset.
However, the experiments were conducted using convolutional neural networks tested with harsh conditions such as poor weight initialization and different activation functions.
Due to long simulation times, we limited the number of ensembles in this case.
Future work will investigate a more variable ensemble size.
We implemented a slight modification of the EnKF in which poorly performing individuals can be replaced by the best individuals.
The fitness is used to rank the individuals and replace the worst $n$ individuals with $m$ best ones.
Furthermore, we add random values drawn from a normal distribution to the replacing individuals in order to increase the search space for the parameters and to find different and possibly better solutions.
We set $n \text{ and } m$ to be $10\%$ of the corresponding individuals.
One hyper-parameter of the EnKF is $\gamma$ (set to $\gamma=0.5$), it can be compared to the effect of the learning rate in stochastic gradient descent.
A lower $\gamma$ may lead to a faster convergence but also has the risk of overshooting minima.
In contrast a higher $\gamma$ is slower to converge, or can get trapped in minima.


\subsubsection{Analysis}\label{subsubsec:uc1_results}
\begin{figure}[ht]
  \centering
  \includegraphics[width=0.6\linewidth]{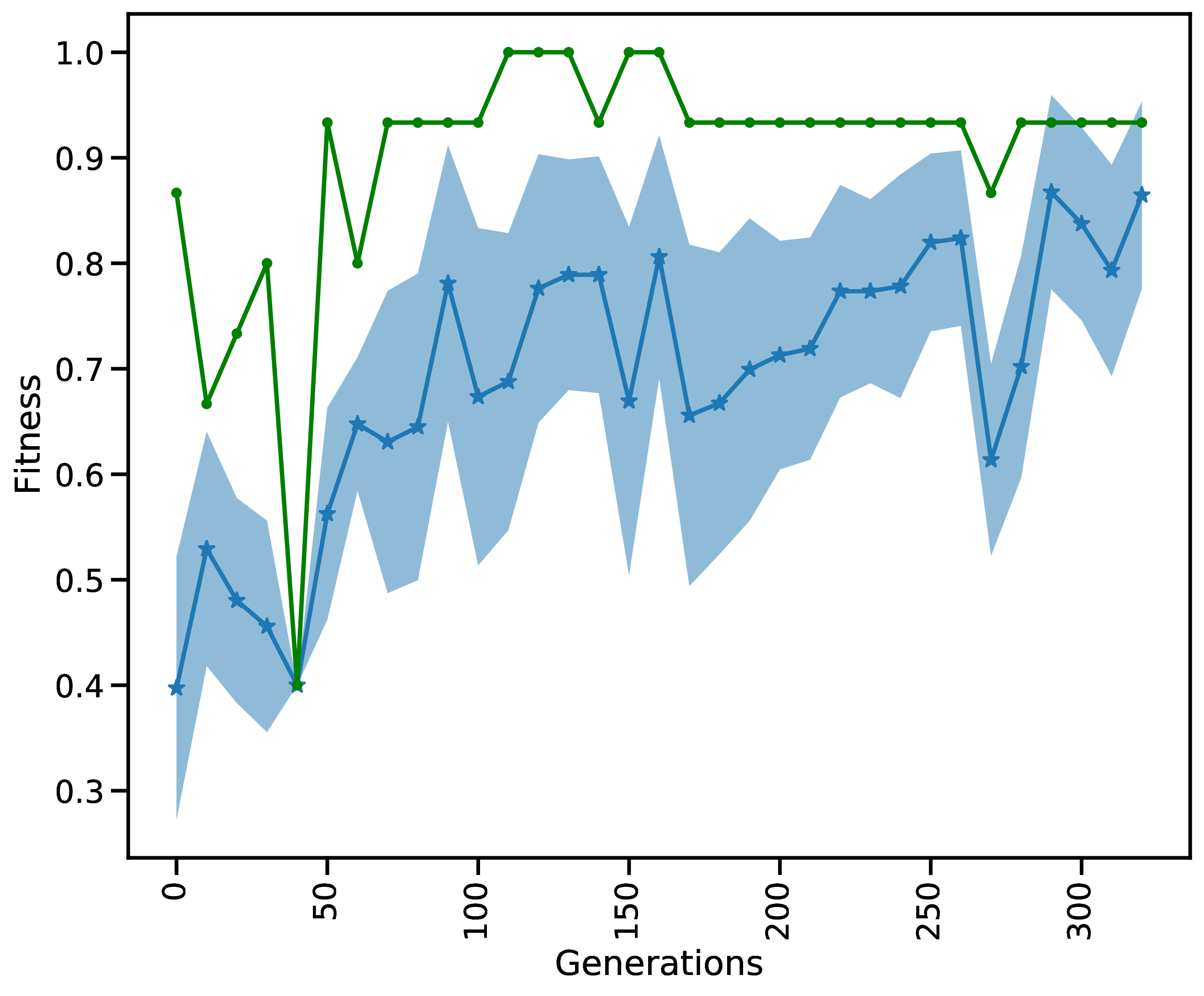}
  \caption{\label{fig:nest_fitness} Every tenth iteration the reservoir is tested on a small part of the MNIST test data. The blue dotted line shows the mean fitness and the shaded area is the standard deviation of all individuals. The green line depicts the best fitness in every generation.}
\end{figure}

Figure~\ref{fig:nest_fitness} depicts the evolution of the fitness over $320$ generations.
The test is acquired over a subset of the MNIST test set in every tenth generation.
The test set is separated from the training set and contains digits that were not presented during training.

While the mean fitness steadily increases over the generations, the best individual fitness exceeds $0.9$ at generation $50$ and improves to a fitness very close to $1.0$ before decreasing again to around $0.9$.
Towards the end of training, we observe that the standard deviation of the individuals gets smaller and the mean increases.
After a maximum standard deviation of $0.16$ in generation $100$, the spread of the ensemble contracts to a minimum standard deviation of $0.08$ in generation $260$, and remains low thereafter.
It is important to note that the green curve indicates the performance of the highest performing individual in each generation, this is not necessarily the same individual.
In this setting we focused explicitly on three digits of the dataset (0 to 2)
Currently we show 10 images for $500\MS$ on each generation which takes relatively long simulation times, thus hindering our ability to process the whole dataset.
Although the simulations take a relatively long time, using the HPC capabilities of L2L we are able to process an entire generation of $98$ individuals including the optimization of a total of $98 \times 28800$ weights in less than 3 minutes.
In comparison a grid search on $28000$ parameters  exploring a range of $20$ values for each weight would require the evaluation of $20^{28000}$ combinations.
Due to the fast convergence behavior of the EnKF it is possible reach an optimal solution in few generations.
Our modifications to sample new individuals from well performing ones and perturbing them increases the possibility to find an overall better solution by exploring other parameter ranges.
A future research direction we want to investigate is to move the optimization process of the weights into the inner loop and optimize the hyper-parameters of the optimizer.
In this light, it would be interesting to use Nengo or Norse which are suitable for solving machine learning tasks with SNNs and optimize the hyper-parameters of the optimizers provided by those libraries.
Finally, we can compare the results by executing the same approach having NEST as the SNN back-end.

\subsection{Use case 2: Fitting electrophysiological data with Arbor}%
\label{subsec:uc2_arbor}
This use case is concerned with optimizing the parameters of a biophysically realistic single cell model implemented
in Arbor such that the response of the neuron to a specific input stimulus matches an experimental recording. Both
passive parameters -- morphology and resistivities -- and active response to an external stimulus are commonly recorded
in electrophysiological experiments. Similarly, the ion channels present are typically known.
However, the internal parameters of the mechanisms -- usually implemented as a set of coupled linear ODEs -- are not
known. To address this, we use L2L to fit the model parameters to the available data.
This proof-of-concept aims at providing a robust way for model fitting for the Arbor simulator using HPC resources.

\subsubsection{Description of the simulation tool}%
\label{subsubsec:uc2_description}
Arbor is a library for writing high-performance distributed simulations of networks of spiking
neuron with detailed morphologies~\citep{arbor:2019}. Arbor implements a modification of the cable-equation model
of neural dynamics which describes the evolution of the membrane potential over time, given the
trans-membrane currents. In this model, neurons comprise a tree of \textit{cables} (the morphology),
a set of dynamics assigned to sub-sections of the morphology (called \textit{ion-channels} or \textit{mechanisms}),
and a similar assignment of bio-physical parameters. The morphology describes the electric connectivity
in the cell's dendrite and the mechanisms primarily produce the trans-membrane currents.

\subsubsection{Optimizee: Morphologically-Detailed Single Cell}%
\label{subsubsec:uc2_task}
\begin{figure}
    \centering
    \includegraphics[width=0.35\textwidth]{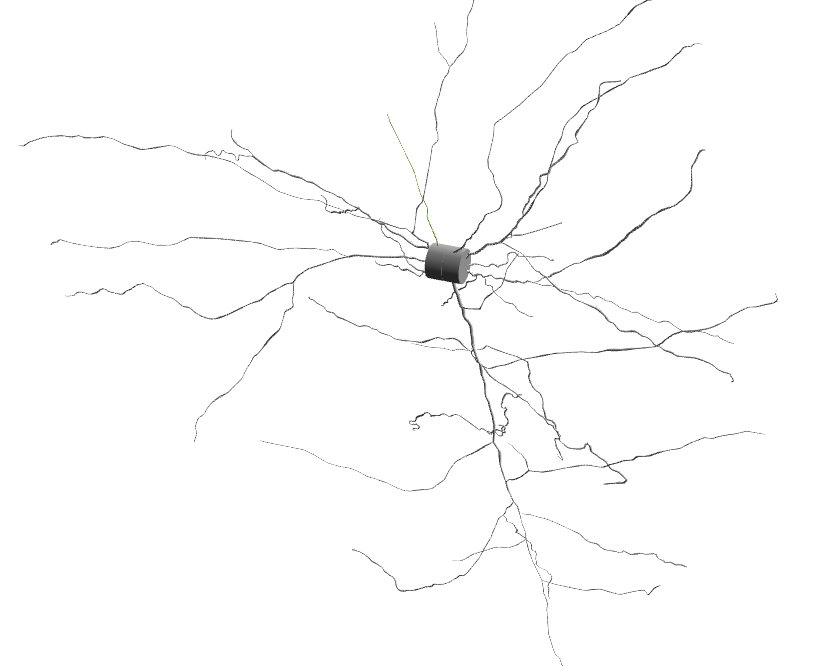}
    \includegraphics[width=0.5\textwidth]{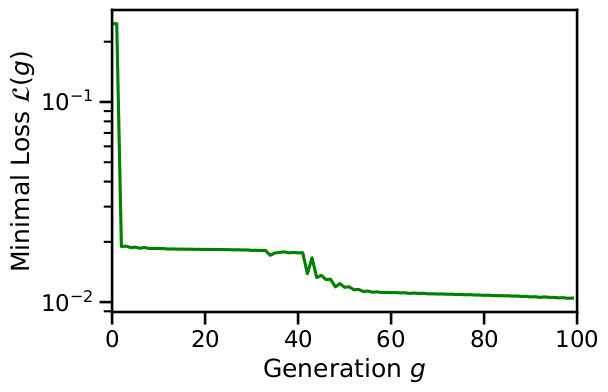}
    \caption{Example input to Arbor and trace of a run of the optimizer.\newline
    (Left) Cell morphology as consumed by Arbor and imported from the Allen DB, regions are marked as `soma',
    `dendrite', and `axon'.\newline
    (Right) Minimal loss function over generations of the genetic optimizer for an example run of L2L on this
    cell starting from random parameters.}%
    \label{fig:arbor-fitness}
\end{figure}

As outlined above, we expect models to be imported from laboratory data, that is a morphological description of the
cell from microscopy, a template of ion channels with yet unknown parameter values, and some known data like the temperature
of the sample. In addition, a series of stimulus and response measurements needs to be provided, which will be the
target of optimization. Our objective then is to assign values to the parameters to best approximate the measured response.
For designing this use-case, we focus on a single specimen from the Allen Cell Database with a known parametrisation
in addition to the input/response data~\cite{allen2007}.

We define the parameter sets $\mathbf{P}$ to be fit as a list of 4-tuples: a sub-section of the morphology,
an ion-channel id, a parameter name, and the value to set the parameter to.
Regions in the morphology are written as queries against Arbor's layout engine, e.g. selecting all parts of the
dendrite where the cable  radius is smaller than $\SI{1}{\micro\metre}$ becomes \verb!(rad-lt (tag 2) 1)!,
since \texttt{tag=2} as been set during morphology creation.
Consequently, setting the parameter \verb!tau! in the \verb!expsyn! mechanism to $\SI{2}{\milli\second}$ appears as
\begin{verbatim}
[.., ((rad-lt (tag 2) 1), expsyn, tau, 2), ..]
\end{verbatim}
in the individual. Optimizee instances are constructed from are configuration file which lists the following items (example item)
\begin{itemize}
    \item morphology file name (\verb!cell.swc!)
    \item list of current clamps (\verb!(delay, duration, amplitude)!)
    \item simulation parameters: length and time-step
    \item location where to record the response (\verb!(location 0 0.5)!)
    \item reference response (\verb!ref.csv!)
    \item fixed parameter assignments (\verb!T=285 K!)
    \item list of ion channel assignments and optimizable parameters
          (\verb!(tag 2), pas, e, -70, -30!)
\end{itemize}
Parameters to be optimized are given a bounding range used to automatically restrict the optimizer, here \verb!e! may vary in the range of $[-\SI{70}{\milli\volt}\dots-\SI{30}{\milli\volt}]$. This data is sufficient -- together with the statically known items -- to
construct a simulation in Arbor that can be run forward in time.

\subsubsection{Fitness metric}%
\label{subsubsec:uc2_metric}
\begin{figure}
    \centering
    \includegraphics[width=0.45\textwidth]{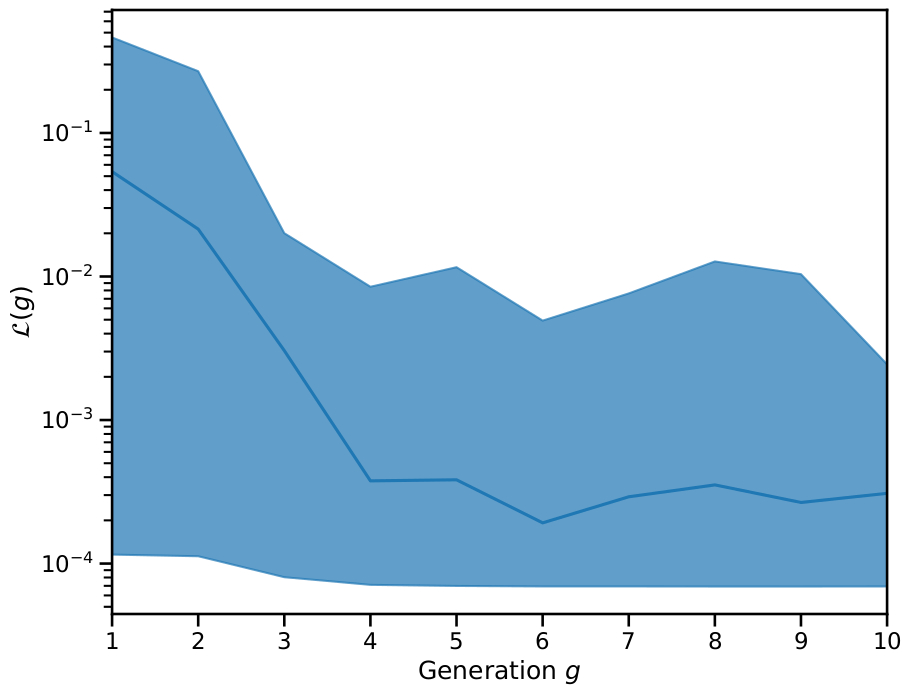}
    \includegraphics[width=0.49\textwidth]{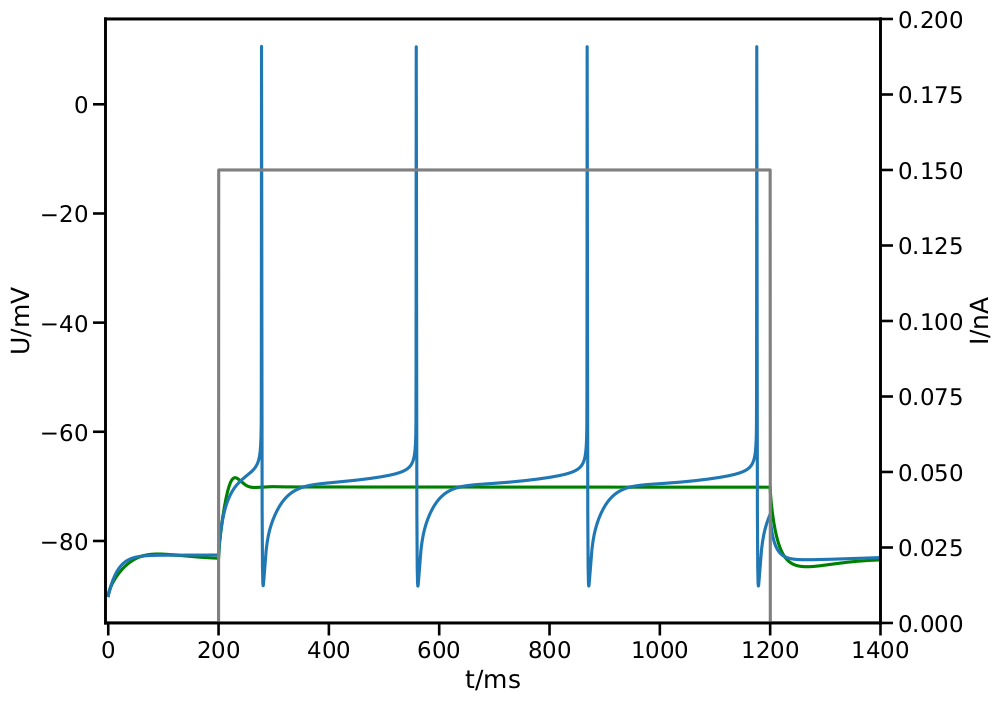}
    \caption{%
    Initial evolution of the optimisation process until generation ten.\newline
    \textbf{Left:} Fitness evolution for the first ten generations, shown are the average, minimum, and maximum loss.\newline
    \textbf{Right:} Best result after $10$ rounds of the optimiser with a loss of $0.0007$.
    Measured membrane potential at the center of the soma from the simulation~$\textcolor{snsgreen}{\bullet}$ against reference~$\textcolor{snsblue}{\bullet}$ and applied stimulus~$\textcolor{black!50}{\bullet}$.
    }%
    \label{fig:arbor-trace}
\end{figure}

We implemented the naive approach of using the mean square loss as the measure for fitness. Given the
experimentally obtained membrane potential $U_{\mathrm{ref}}(t)$ we define the fitness as
\begin{align}
    \mathcal{L}(\mathbf{P}) = -\frac{1}{T^2}\sum\limits_{t=0}^{T}\left[U_{\mathrm{ref}}(t\cdot \tau) - U_{\mathrm{sim}}(\mathbf{P}, t\cdot \tau)\right]^2
\end{align}
where $ V_{\mathrm{sim}}(\mathbf{P}, t)$ is the measurement produced by Arbor given the parameter set $\mathbf{P}$
and $\tau$ is the sampling interval of the voltage measurement. The optimizer attempts to maximise the given metric,
which is why we defined the fitness as the negative of the $L_2$ norm here.

Figure~\ref{fig:arbor-fitness} shows an example of a single cell morphology and the loss function across a single
run of L2L. As can be seen in Figure~\ref{fig:arbor-trace}, we quite easily reach a configuration that
reproduces the \emph{mean} membrane voltage, but not does exhibit spiking behaviour. From experience we know that spikes
are only produced for a narrow band of parameters in these complex configurations.

Thus, the fitness function will need to be
extended to include the requirement for spiking. Further it seems prudent that the final result of the optimisation process
should include the responses to multiple separate stimulation protocols. Therefore, the overall fitness becomes a vector
\begin{align}
\mathcal{F}(\mathbf{P}, \mathbf{I}) = \left(
\begin{matrix}
 \mathcal{L}(\mathbf{P}, I_0)\\
 \mathcal{S}(\mathbf{P}, I_0)\\
 \mathcal{L}(\mathbf{P}, I_1)\\
 \vdots\\
\end{matrix}
\right)
\end{align}
which -- in conjunction with a vector of weights  -- is suited for use with L2L's
multi-objective optimisation.
Here $\mathbf{I}$ is the vector of stimuli and the function $\mathcal{S}$ collects the fitness with respect to the spiking behaviour.
The simplest choice for $\mathcal{S}(\mathbf{P}, I)$ is the overall spike frequency. However, using spiking behaviour as a metric
is in itself problematic: As said before, spikes only incur in a narrow band of parameters, thus resulting in a fitness function
with essentially constant zero score except in this narrow band where the score is close to its maximum. This poses problems for
optimizers relying on smooth variation in $\mathcal{F}$ to find the target parameter set.

\subsubsection{Optimizer: Evolutionary algorithm}\label{subsubsec:uc2_outer}
The fitness metric is used to drive the outer loop optimizer, an evolutionary algorithm searching
for the maximum fitness. This class of algorithms has been proven as computationally efficient for this kind of fitting problems~\citep{druckmann2007novel}.

In the L2L framework, the Genetic Algorithm optimizer (GA) is a wrapper around the DEAP library~\citep{DEAP_JMLR2012}.
This adapter takes care of handling the parameters received from the inner loop and prepares them for the optimization process.
The DEAP library then facilitates the cross-over and mutation methods, applies them on the actual parameter set and saves the best individuals into the Hall of Fame if they fare better than previous runs.
Afterwards, the optimized parameters are sent back to the optimizee, which then initializes the next generation of individuals.

Here, we use a population of 100 individuals and a total of 200 generations. Individuals in a generation are evaluated in using
16 parallel tasks on a single dual-socket node.

\subsubsection{Analysis}%
\label{subsubsec:uc2_results}
We have shown a basic implementation for finding optimal parameter sets for single cell models using Arbor and
L2L. This enables researchers to fit experimental data to neuron models in Arbor, a workflow that is important
in practice and lacking so far in Arbor's ecosystem. The approach shown here so far is implemented in a straightforward
fashion, but falls short to reach the desired configuration in a reasonable time frame. As discussed above, extending the
fitness function to a more sophisticated implementation will be addressed to alleviate this problem.
It might also be profitable to rely on salient features instead of using the membrane potential directly~\citep{druckmann2007novel, gouwens2018systematic}. To cope with common time-restrictions on the used resources in the mean-time we implemented a method
to resume optimization given an intermediate result. Currently, this workflow is being extended beyond the proof-of-concept state we presented here.

Another extension is the use of accelerators (GPUs), which allow for massively parallel evaluation of individuals.
Arbor is able to use GPUs for simulations efficiently starting at a few thousands of cells per GPU. This would enable
processing an entire generation of the optimization process at once. Given the current numbers of 100 cells per generation
this is not yet profitable, but for larger generation sizes and additional stimulus protocols it becomes attractive. L2L
was extended to enable a vectorized version of the evolutionary algorithm similar to the multi-gradient
descent approach used in use case \ref{subsec:uc4_tvb}.

\subsection{Use case 3: Foraging behaviour with Netlogo and NEST or SpikingLab}\label{subsec:uc3_netlogo}

\begin{figure}[ht]
  \centering
  \includegraphics[width=0.37\linewidth]{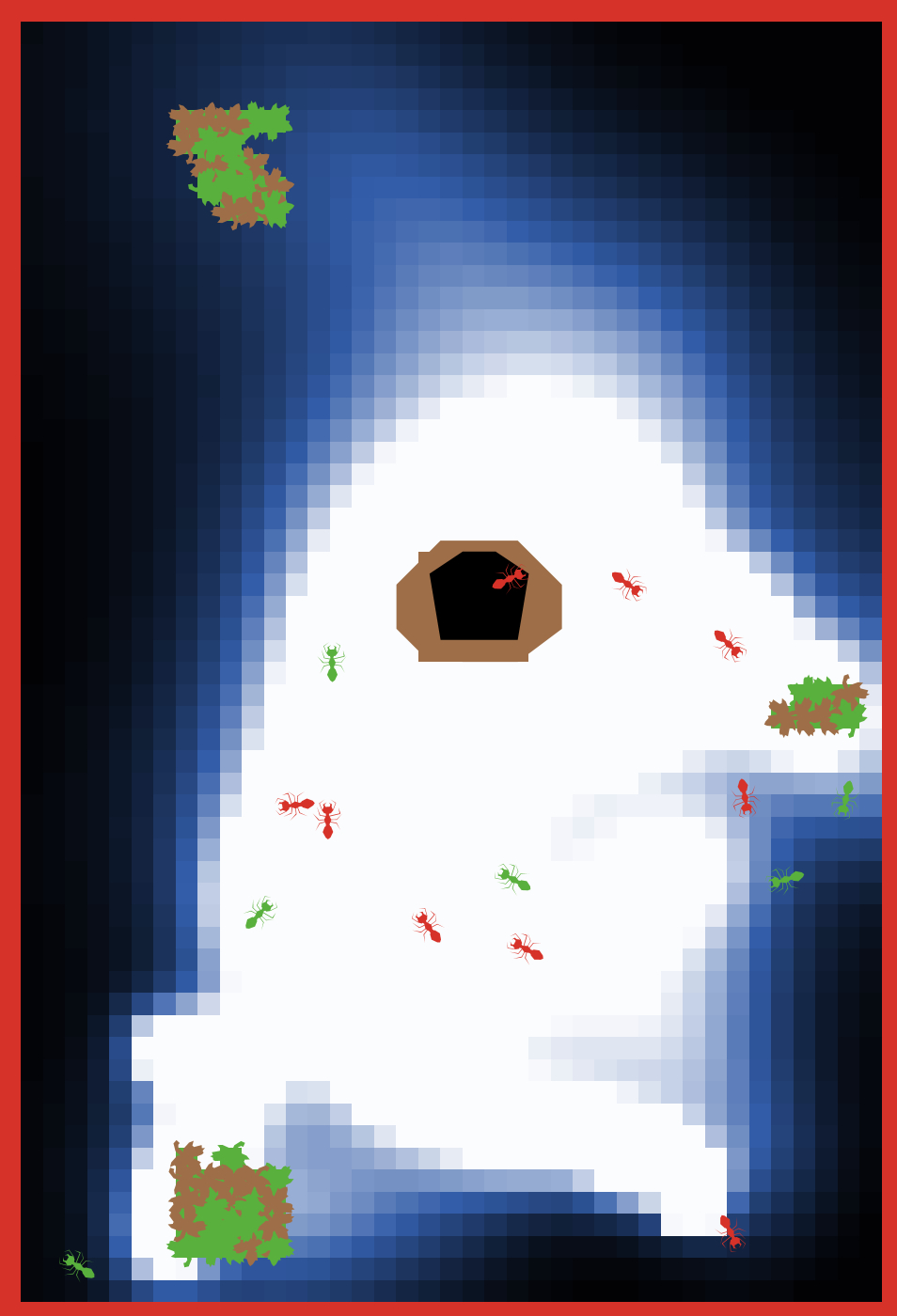}
  \caption{\label{fig:ant_colony} The ant colony is searching for food (big green patches with brown leaves). The ants are communicating via pheromones which are dropped on the ground (blue-white patches) when food is found or when the ants return to the nest (black-brown patch). Green colored ants are transporting the food, while orange colored ants are exploring the environment or following the pheromone trail. The red border around the world is a impenetrable wall and prevents ants crossing from one side to the other. The pheromone trail decays with time if it is not reinforced by other ants.}
\end{figure}

In this use case we describe optimizing the foraging behaviour in a simulated ant colony.
The colony consists of 15 ants, all of which are searching for food  (big green patches, Figure~\ref{fig:ant_colony}).
Any food found must be brought back to the nest.
Ants communicate with each other by dropping pheromones on the ground (blue patches) whenever the food is found or the nest is reached.
The pheromone can be smelled by other ants which then can follow the trail left on the ground.
Each ant is controlled by a spiking neural network, which is an identical copy for every ant.
Here, we use L2L to configure its weights and delays so that the ants bring food back to the nest as efficiently as possible.

\subsubsection{Description of the simulation tools}\label{subsubsec:uc3_description}
NetLogo is a multi-agent simulator and modeling environment~\citep{tisue2004netlogo}.
It is widely used as an educational and scientific tool for the study of emergent behavior in complex systems.
Agents are expressed as objects that can communicate with each other.
In our setting, NetLogo helps us to observe and manipulate the state of every neuron and synapse.
For the simulations we have two backends: NEST (see Section~\ref{subsubsec:uc1_description}) and SpikingLab~\citep{jimenez2017spikinglab}.
SpikingLab is an engine directly integrated within NetLogo and can be easily and quickly used for small scale networks, as we present in our use case. Invoking NEST from NetLogo causes a minimal communication overhead since NEST needs to be called as an external process. For larger networks, the greater efficiency of NEST more than compensated for this overhead, and it becomes the preferred option.

\subsubsection{Optimizee: Simulated ant brain}\label{subsubsec:uc3_inner}
\begin{figure}
    \centering
    \includegraphics[width=0.8\linewidth]{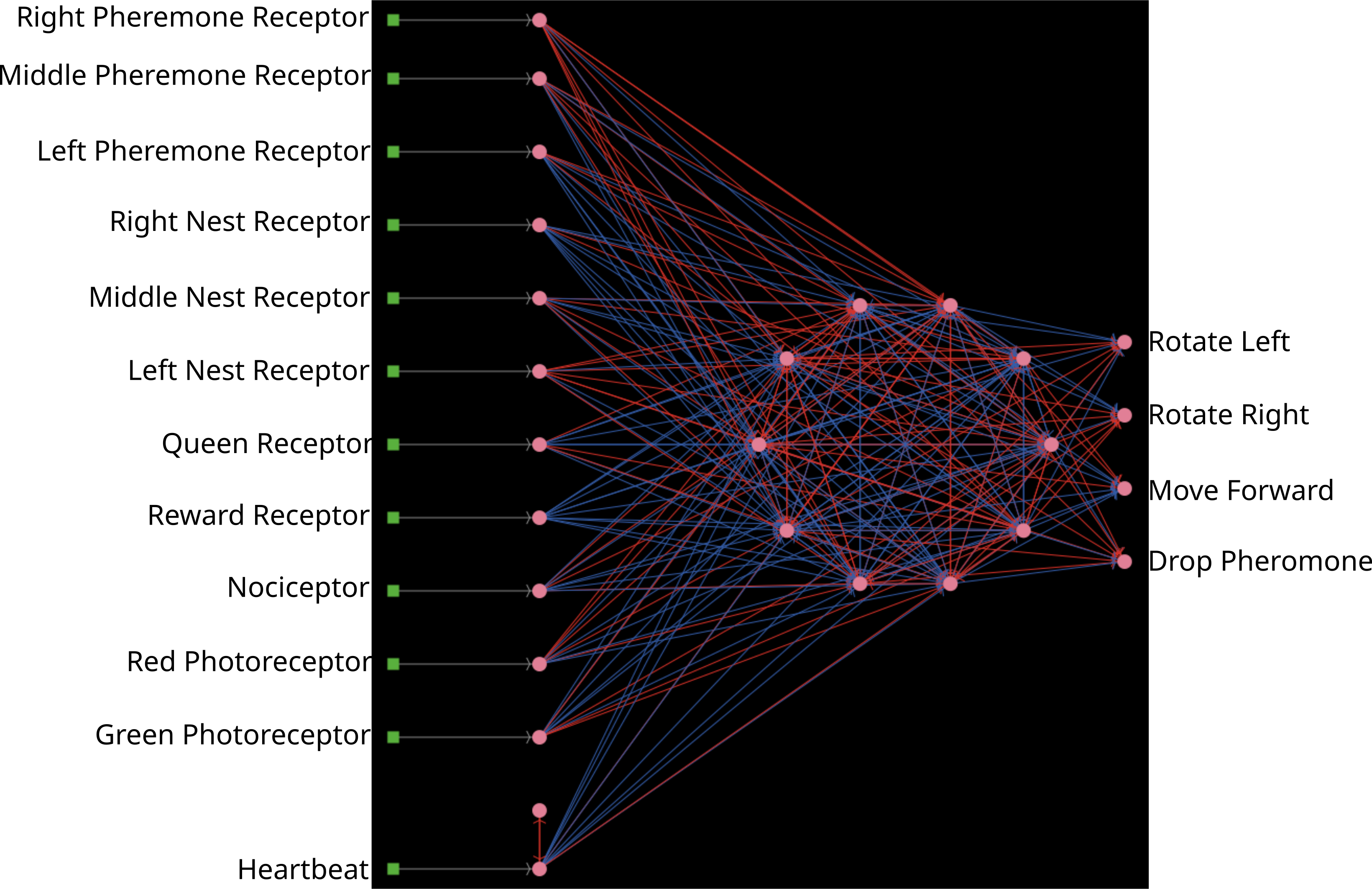}
    \caption{The SNN for the ant colony. Every ant is steered by an SNN. Neurons are depicted as pink dots and excitatory/inhibitory connections as red/blue lines. All networks are identical.}
    \label{fig:colony_net}
\end{figure}
In the first iteration the optimizee creates the individual inside the \texttt{create\_individual} function.
The individual consists of network weights and delays.
The weights ($220$) are uniformly distributed in $[-20, 20]$, while the delays ($220$) range between $[1, \ldots, 7)_{\mathbb{N^{+}}}$.
The network has an input, a hidden and an output layer, the neurons are all-to-all connected for every layer as depicted in Figure~\ref{fig:colony_net}.
The input layer consists of $12$ neurons.
The first three neurons are receptors to smell the direction of the pheromone.
The next three neurons are responsible to locate the nest.
The queen receptor indicates the middle of the nest.
Reward and nociceptors determine the reward and punishment for the ant.
The green and red photoreceptor are triggered when food or a wall is seen.
Finally, the heartbeat neuron stimulates the network in every timestep with a small direct current to keep a low dynamic ongoing in the network.
The four output neurons are responsible for the movement and for dropping the pheromone.
Similar to the first use case in Section~\ref{subsubsec:uc1_results} the total number of individuals is $98$.
The weights and delays can be min-max normalized if specified.
The optimizee saves these parameters as a csv file before starting the simulation.
The model is invoked by a Python subprocess~\footnote{https://docs.python.org/3/library/subprocess.html} in the \texttt{simulate} function, which then calls the headless mode of NetLogo to start the run.
The optimizee waits until the simulation is finished and collects the fitness value from a resulting csv file which
is written after the simulation ends.

The user has to set whether NEST or SpikingLab is invoked as a backend inside the simulation.
NEST is called as a subprocess by NetLogo, while SpikingLab is directly accessed by the model.
In the case that NEST is selected, the parameters have to be passed to it as well, since the network needs to be constructed with the new parameters.
This can be done either by loading the parameter in a csv file within NEST, or NetLogo can read the csv file and pass the values to the simulation.

The parameters are restricted within the $\texttt{bounding\_func}$ function if their values exceed the specified ranges after the optimization process.
Weights are clipped to the range of $[-20, 20]$ and delays to $[1, 5]$.

\subsubsection{Fitness metric}\label{subsubsec:uc3_metric}
The fitness function for the ant colony optimization problem rewards finding food and bringing it back to the nest whilst punishing excessive movement.
We define the ant colony fitness $f_{i}$ of optimizee $i$ as:
\begin{equation}
 f_{i} = \sum_{t=1}^{T} \left(\sum_{j=1}^{J} \mathcal{N}^{(t)}_{i,j} + \mathcal{F}^{(t)}_{i,j} - \mathcal{C}^{(t)}_{i,j}   \right) ,
 \label{eq:fitness_antcolony}
\end{equation}
where $t = 1, \ldots, T$ is the simulation step and $T$ is the total simulation time, $J$ is the total number of ants in the colony and  $j$ indexes the ants.
$\mathcal{N}$ is the reward for coming back to the nest with food, $\mathcal{F}$ is a reward for touching the food and $\mathcal{C}$ is the movement cost.
Every movement, rotation and pheromone dropping is added towards $\mathcal{C}$.
We set the cost as follows: Rotation $-0.02$, pheromone dropping $-0.05$, movement $-0.25$.
The movement has a higher cost, since we would like to restriction vast movements and force to return to the nest.
We also punished resting with $-0.5$ to speed up the movement and to slightly induce exploration.
The rewards are: Returning to the nest $220$ and touching food $1.5$.
A high reward for coming back to the nest is necessary, otherwise the ants are spending a long time exploring the environment even when the food is found.
This slows down learning and hinders solving the task.

\subsubsection{Optimizer: Genetic Algorithm}\label{subsubsec:uc3_outer}
We use a genetic algorithm to optimize the weights and delays in the ant brain network.
This is the same class of optimizer as used in section~\ref{subsec:uc2_arbor}.

\subsubsection{Analysis}\label{subsubsec:uc3_results}
\begin{figure}[ht]
  \centering
  \includegraphics[width=0.6\linewidth]{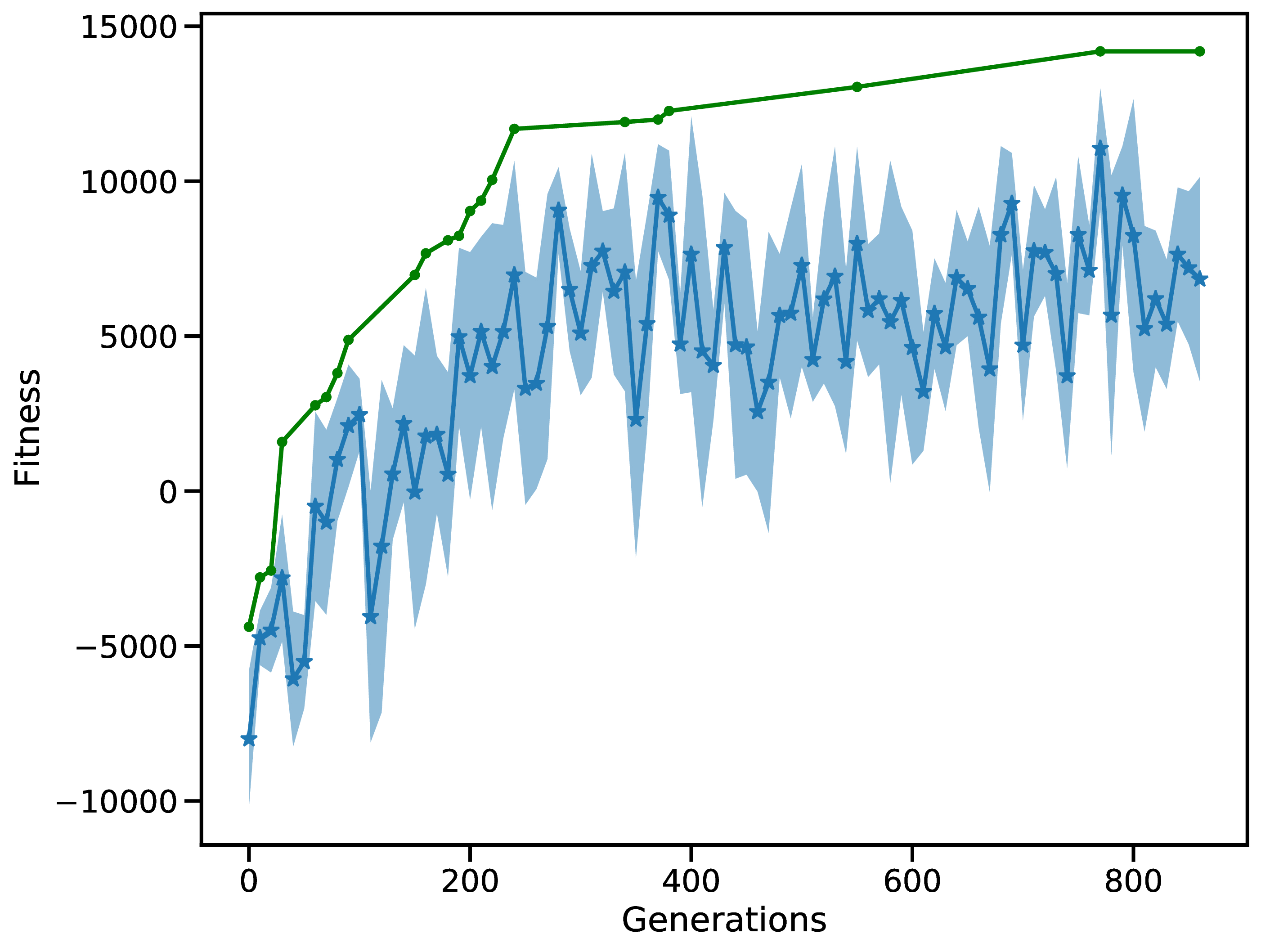}
  \caption{\label{fig:fitness_antcolony} Fitness of the ant colony. The blue curve shows the mean fitness and the shaded area is the standard deviation. The green curve indicates the best solution found so far, and thus rises monotonically.}
\end{figure}
Figure~\ref{fig:ant_colony} depicts the evolution of the fitness of the ant colony over $800$ generations. Initially, the ants move a lot without retrieving food, resulting in a negative maximum fitness.
After around $200$ generations the mean fitness is consistently positive and the best solution is close to $10000$.
In following generations the mean fitness saturates at around $5000$, with increasing best fitness.
After $800$ generations the L2L run is stopped with the best individual fitness close to $15000$.
Similarly to use case~\ref{subsec:uc1_nest}, L2L enables us to execute $98$ individuals in parallel, where a generation is optimized in less than 2 minutes.
A grid search algorithm with $20$ values to explore weight and delay combinations would require $20^420$ possibilities to test for.
The mutation and cross-over steps of the GA increases the parameter space and avoid local minima, without loosing performance.
The best individuals are saved in the Hall of Fame (HoF) if they have a better fitness than their predecessors.
If an optimization step produces underperforming individuals it is possible to recombine the new set utilizing the HoF.
Due to the parallel distribution of individuals and the GA optimizer we are able to find well performing individuals in less than $400$ generations.

\subsection{Use case 4: Fitting functional connectivity with TVB}\label{subsec:uc4_tvb}
This use case describes tuning the parameters of a whole brain simulation using the GPU models of The Virtual Brain simulator ~\citep[TVB;][]{sanz2013virtual} to give the best match to empirical structural data.

To do clinical research with TVB it is often necessary to configure the parameters of a model for a specific person such that it matches obtained empirical data.
First, the brain is parcellated into different regions, based on many available atlases~\citep{bansal2018personalized}.
The connectivity of these regions is determined using diffusion weighted imaging, estimating the density of white matter tracts between the regions, resulting in a connectivity matrix which is regarded as the structural connectivity.
Finally, a model that represents the regional brain activity must be chosen.
To optimize the match between a specific person and the TVB simulation, obtained fMRI can be used to further personalize the structural connectivity~\citep{deco2014identification}.

Due to the high dimensionality of TVB models and the wide variation in possible parameter values, fitting patient data often requires extensive parameter explorations over large ranges.
In this use case the simulated functional connectivity is matched to the structural connectivity.
The task has the underlying assumption that regions which are anatomically connected often show a functional connection \citep{Honey2035}.
In this task we want to find the values for the \texttt{global\_coupling} and \texttt{global\_speed} variables, characteristic to the connectome of a TVB stimulation, which give rise to the strongest correlation between the structure of the brain and the functional connectivity, i.e.~the relationship between spatially separated brain regions.


\subsubsection{Description of the simulation tools}\label{subsubsec:uc4_description}

The Virtual Brain is a simulation tool which enables researchers to capture brain activity at mesoscopic level using different modalities such as EEG, MEG of fMRI, using realistic biological connectivity.
A TVB brain network consists of coupled neural mass models (NMM) whose dynamics can be expressed by a single or system of ordinary differential equations.
The coupling of the NMMs is defined by the connectivity matrix.
The NMMs describe, e.g.~the membrane potential or firing rate of groups of neurons, represented by differential equations which are solved numerically, in this case, with an Euler based solver.
RateML~\citep{rml}, the model generator of TVB, enables us to create the desired TVB model written in CUDA for the GPU and a driver to simulate the model, from a high level model XML file.


\begin{figure}[ht]
  \centering
  \includegraphics[width=0.6\linewidth]{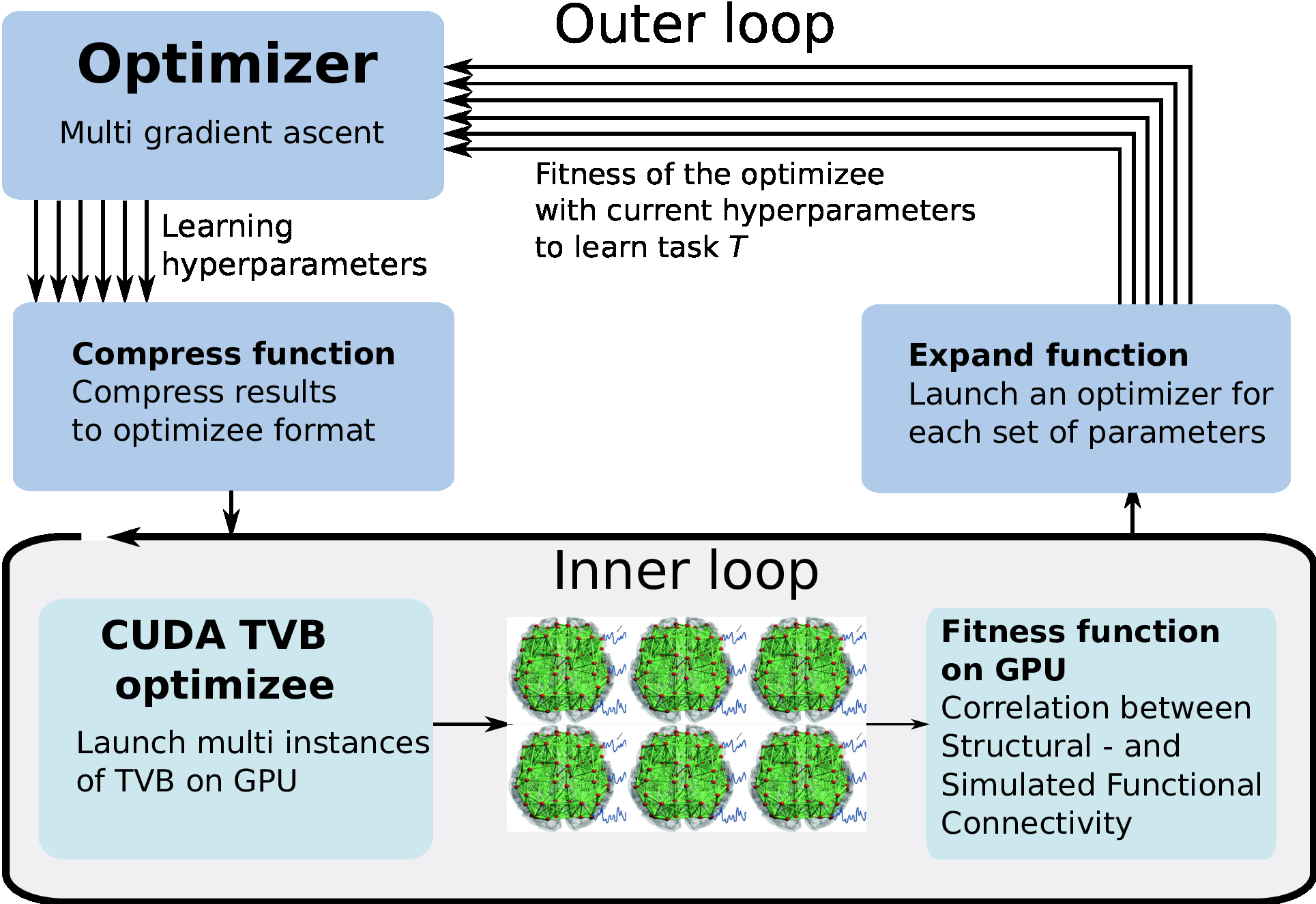}
  \caption{\label{fig:multiagent} The multi-gradient ascent implementation of L2L.
  The inner loop launches multiple instances of TVB on the GPU simulating different sets of parameters.
  The outer loop select the best fitnesses and produces a new parameters range.}
\end{figure}
Unlike the use cases discussed above, in this case we exploit GPU-parallelization by defining an optimizer that can process a vector of fitnesses, and create new individuals for multiple TVB simulations executed in parallel on the GPU.
An overview of this process is shown in Figure~\ref{fig:multiagent}.
The optimizee in the inner loop spawns a number of threads (here: $1,024$) according to the users defined parameters ranges and resolution.
Each thread represents a TVB instance, simulating a unique set of parameters. The fitness is computed for each instance, and the outer loop optimizer selects the best fitness by using the gradient ascent strategy.
The arrows indicate the independent iterations of the vector of fitnesses.
In the figure, six TVB simulations run in parallel, thus the optimizer needs to iterate a vector of six fitnesses.

\subsubsection{Optimizee: Whole brain simulation}\label{subsubsec:uc4_inner}
The \texttt{create\_individual} function initializes a first instance for the TVB simulation. The structural connectivity is usually obtained from the patient but in this case the standard TVB connectivity for $76$ nodes is used. We model the regions with the \texttt{Generic2DimensionOscillator}~\citep[G2DO;][]{Ott2008}. A dictionary is created which contains initial random values for the optimization parameters,  \texttt{connection\_speed} and \texttt{coupling\_strength}.

For subsequent simulation generations, the optimizee reads the adapted values from a text file written by the optimizer and utilizes the Python subprocess module to spawn a new TVB simulator object with the corresponding parameterization.
When the TVB simulation is complete, the fitness for each TVB instance is computed and written to a separate text file.
The text files are read by the optimizee reformatted for processing by the optimizer.

\subsubsection{Fitness Metric}\label{subsubsec:uc4_metric}
The computation of the fitness for this task is two fold.
In a first step the simulated functional connectivity is determined by computing the Pearson product-moment correlation coefficient, $\rho_{xy}$, of the simulated $76$ regions according to Equation~\ref{eq:pearsson}.

\begin{equation}
 \rho_{xy} = \frac{\mathrm{Cov}(x,y)}{\sigma_{x}\sigma_{y}} \, ,
 \label{eq:pearsson}
\end{equation}

Where $\mathrm{Cov}(x,y)$ is the covariance of variables $x$ and $y$ and $\sigma_{x}$ and $\sigma_{y}$ is the standard deviation.
This first step determines how strong the dynamics of the simulated regions correspond to one another.
A strong functional correlation means that the simulated activity between the spatially separated brain regions is more similar.
The second step is to determine the correlation between the obtained functional- and the structural connectivity, the weight matrix used in the simulation, also using Equation~\ref{eq:pearsson}.

\begin{lstlisting}[language=Python,label={code:tvb_coercoef},captionpos=b, caption={Implementation of the correlation computation between functional- and structural connectivity.}]
SC = connectivity.weights / connectivity.weights.max()
for i in range(couplings * speeds):
	FCSC[i] = np.corrcoef(FC[:, :, i].ravel(), SC.ravel())[0, 1]

\end{lstlisting}
The Python implementation of the second step is shown in Listing~\ref{code:tvb_coercoef}, where \texttt{SC} is the structural connectivity and \texttt{FC} is the simulated functional connectivity which was computed previously.
On line 1 the weights are normalized.
In the for-loop on line 2, the correlation with the structural connectivity is computed.
The \texttt{FCSC} holds these correlations and is the array of fitnesses returned to the optimizer.


\subsubsection{Optimizer: Multi-gradient ascent}\label{subsubsec:uc4_outer}
The best fitness is selected with a gradient ascent optimizer.
The existing optimizer has been adapted for processing the vector of fitnesses returned by the GPU, named multi-gradient ascent (MGA).
In order to adapt it to vector processing the fitnesses need to be expanded before processing and compressed afterwards, as is shown in Figure \ref{fig:multiagent}.
The expanding transforms the obtained fitnesses from the optimizee process to a data structure in which the obtained fitnesses are linked to the used parameters, thus enabling the multi-gradient ascent optimizer the possibility to select the best fitness and define a range for the new parameters to be sent to the optimizee.
When the optimizer has selected the parameters for the optimizee, it compresses the new individuals to a data structure that just contains the new parameter combinations for the optimizee.
Aside from the expanding and compressing, the MGA algorithm determines the new values for the individuals similar to gradient ascent.

\subsubsection{Analysis}\label{subsubsec:uc4_results}

\begin{figure}[ht]
  \centering
  \includegraphics[width=0.6\linewidth]{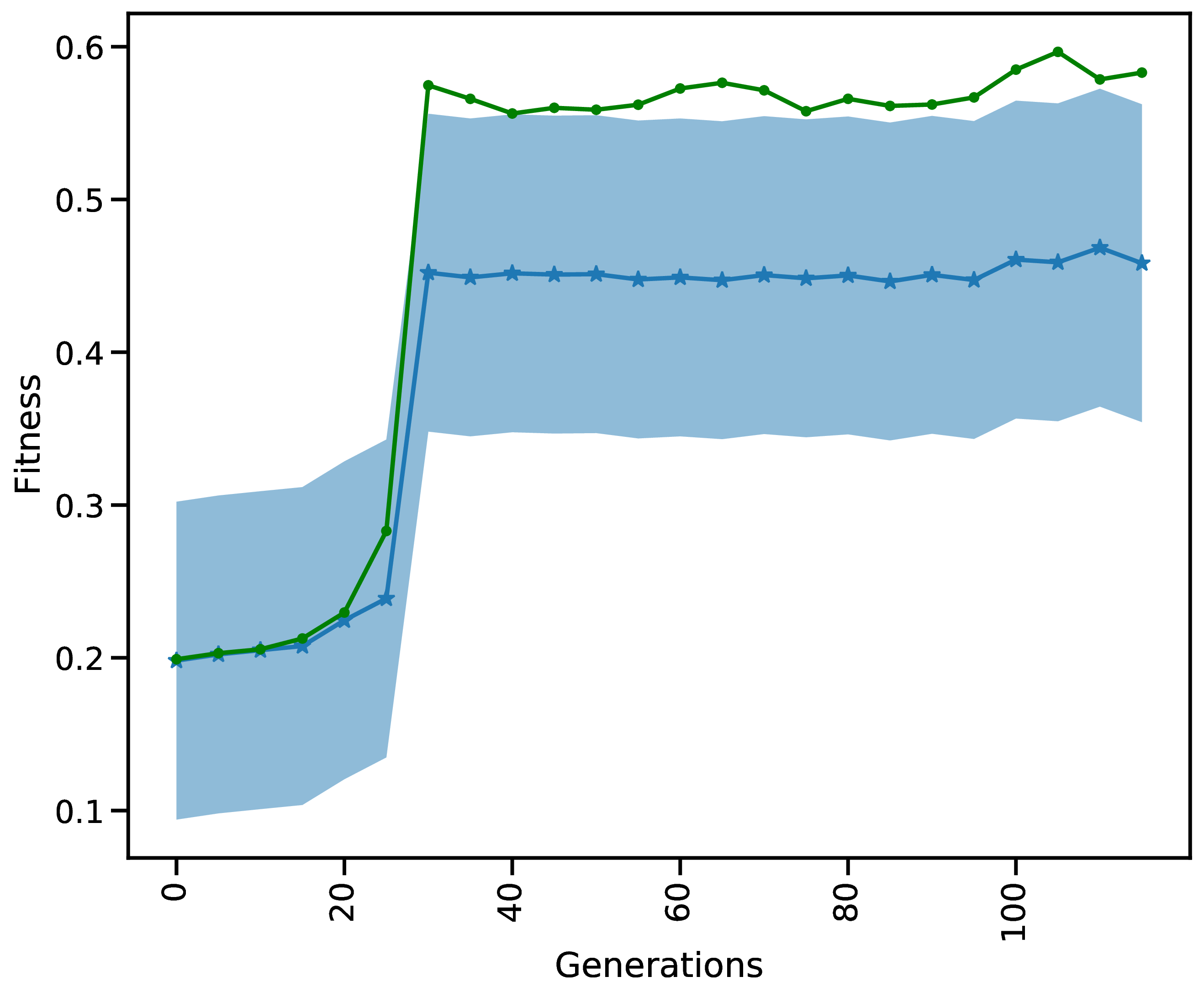}
  \caption{\label{fig:multigrad} Standard deviation of the mean and best out of $1,024$ fitnesses for $116$ generations for the multi-gradient TVB hyper-parameter optimization TVB simulation.
  The blue curve is the mean fitness over the population of $1,024$ and the shaded area gives the standard deviation. The green curve shows the best fitness for each generation.}
\end{figure}

The results in Figure~\ref{fig:multigrad} shows the evolution of the mean and best fitness for a generation of $1,024$ parameter combinations for the \texttt{global\_speed} and \texttt{global\_coupling} variables, with a learning rate of $0.01$ and four individuals.
These four individuals each spawn $1024$ TVB simulations on the GPU, enlarging the chance of success.
Each generation contains a TVB simulation of $4,000$ simulation steps with a $dt = 0.1$.
These results were obtained using a NVIDIA V100 GPU on the JUSUF~\footnote{\url{https://fz-juelich.de/ias/jsc/EN/Expertise/Supercomputers/JUSUF/JUSUF_node.html}} cluster.
Our results show that after $30$ generations the best attainable fitness (green curve) is reached (c.f.~\cite{deco2014identification}).

Comparing the GPU population based to a single L2L implementation, the latter would need more generations before the best fitness is attained.
The likelihood of finding a suitable solution in earlier generations rises with the size of the population:
the more configurations considered in a single generation, the faster it converges to the best value.
The GPU implementation has already considered $30\times 1,024$ different parameters values, after which the optimal fitness is found (Figure \ref{fig:multigrad}), while the single implementation would have only $30$.
A single implementation would need at least $30720$ generations to find the same result, but would very likely need many more.
Additionally, the GPU makes it very convenient to execute many simulations in parallel by not having to split them up onto multiple nodes, without communication overhead and decreasing wall clock time even further.

\subsection{Use case 5: Solving the Mountain Car Task with OpenAI Gym and NEST}\label{subsec:uc5_gym}
\begin{figure}[ht]
  \centering
  \includegraphics[width=1.0\linewidth]{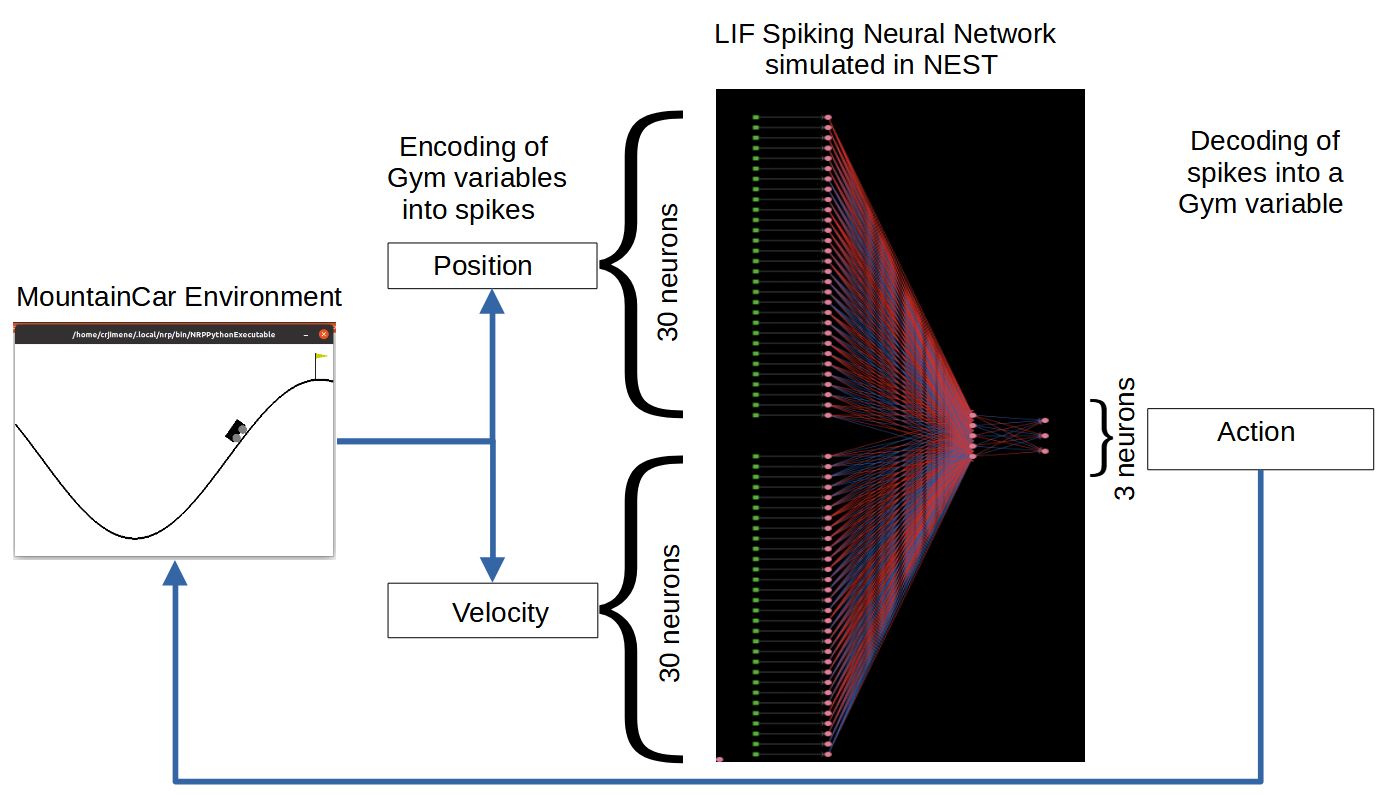}
  \caption{\label{fig:mc_nest} A feed-forward spiking network to solve the Mountain Car task. In the Mountain Car environment (left) the agent must steer the car to reach the goal position (flag on top of the hill). The position and velocity of the car reported by the MC are encoded into spikes by the encoding layer (DC generators depicted as green dots) of the three-layer spiking neural network (neurons depicted as pink dots and excitatory/inhibitory connections as red/blue lines) running in the NEST simulator (right). The activity from the three output neurons (accelerate left, accelerate right, do nothing) is decoded into actions for MC. The set of weights between the sixty encoding neurons and the three output neurons is the object of optimization.}
\end{figure}

In this use case we describe a solution to the OpenAI Gym Mountain Car (MC) problem. The MC task is interesting since it requires the agent to find a policy in a continuous state space constituted by the position and velocity of the car. At the same time, the action space is discrete, limited to three possible actions: accelerate left, accelerate right, do nothing.
The initial position and velocity of the car is set randomly by the environment; the aim is to reach the goal position (yellow flag) as depicted in Figure ~\ref{fig:mc_nest}. As the car's motor is weak, consistently reaching the goal at the top of the hill requires the agent to learn a policy that swings the car back and forth in order to build up momentum. The challenge is considered solved if the car reaches the goal position in an average of $110$ steps over $100$ consecutive trials. We implement a feed-forward LIF spiking neural network in NEST to encode a policy and optimise the weights so as to improve the ability of the network to solve the task.

\subsubsection{Description of the simulation tools}\label{subsubsec:uc5_description}
The OpenAI Gym~\citep{brockman2016openai} is a software library that provides an interface to a wide range of environments for experimentation with reinforcement learning techniques. NEST has been described in Section~\ref{subsec:uc1_nest}.
Both simulators are instantiated and invoked by the optimizee process which implements the closed-loop interactions. These interactions are synchronized in such a way that for each simulation step of the MC environment the SNN is simulated for an interval of 20ms in NEST. On completion of a simulation interval the state of the network is sampled and fed back as an action to the MC environment.

\subsubsection{Optimizee: Spiking feed-forward policy network}\label{subsubsec:uc5_inner}
The spiking neural network of LIF neurons that controls the actions of the car is implemented in NEST. The inputs to the SNN are the position $[-1.2, 0.6]$ and velocity [-0.7, 0.7] variables which are discretized and encoded using $30$ input neurons for each variable. For the discretization (binning) of the continuous variables the width $(w)$ of the bins is given by the minimum $(min)$ and maximum $(max)$ value of the interval divided by the number of input neurons $(n)$ available for each variable. Each value within the range is discretized into a bin which corresponds to one input neuron:
\begin{equation}
 w = \frac{min + max}{n}
\end{equation}
Once a value falls into a bin, its corresponding neuron is activated by a dc current as provided by a connected dc generator resulting in a firing rate of 500Hz. The $60$ encoding neurons have all-to-all connections to an intermediate layer of five neurons, which in turn have all-to-all connections to the three neurons in the output layer corresponding to the three possible actions.
The action sent to the OpenAI Gym environment depends on the activity of the three neurons in the output (third) layer. Each output neuron represents one of the possible actions. Following a winner-takes-all approach, the neuron with the highest spiking activity determines which action is sent to the OpenAI Gym environment.  Figure~\ref{fig:mc_nest} illustrates the spiking network and the closed-loop interaction with the MC environment on the basis of input variables and output actions.

Similarly to the Netlogo use case (see Section~\ref{subsec:uc3_netlogo}), at the beginning the optimizee creates the individual inside the \texttt{create\_individual()} function.The total number of individuals per generation is 32.
Each individual consists of network weights, which are initially uniformly distributed in $[-20, 20]$. There are $315$ weights corresponding to the $(60\times 5)+(5 \times 3) = 315$  synaptic connections in the network.
The instantiation and orchestration of OpenAI Gym and NEST simulator (including the set-up of the SNN) is carried out by the optimizee. Each simulation runs for $110$ simulation steps (where a simulation step corresponds to an action being sent to the environment) or until the goal position is reached.
Once the simulation is completed the optimizee returns the calculated fitness value to the optimizer.
The $\texttt{bounding\_func()}$ function ensures the weights are clipped to the range $[-20, 20]$ if the values exceed this range after the optimization process.

\subsubsection{Fitness metric}\label{subsubsec:uc5_metric}
The fitness function for the MC optimization problem is defined as the maximum horizontal position reached by the car during an episode comprised of 110 simulation steps, i.e.

$f = \mathrm{max}_T(\vec{\mathcal{P_T}})$

Where $\mathrm{max}_T$ returns the item with the highest value in a vector and $\vec{\mathcal{P_T}}$ contains the position of the car on each simulation step up to $T=110$.
\begin{figure}[ht]
  \centering
  \includegraphics[width=0.6\linewidth]{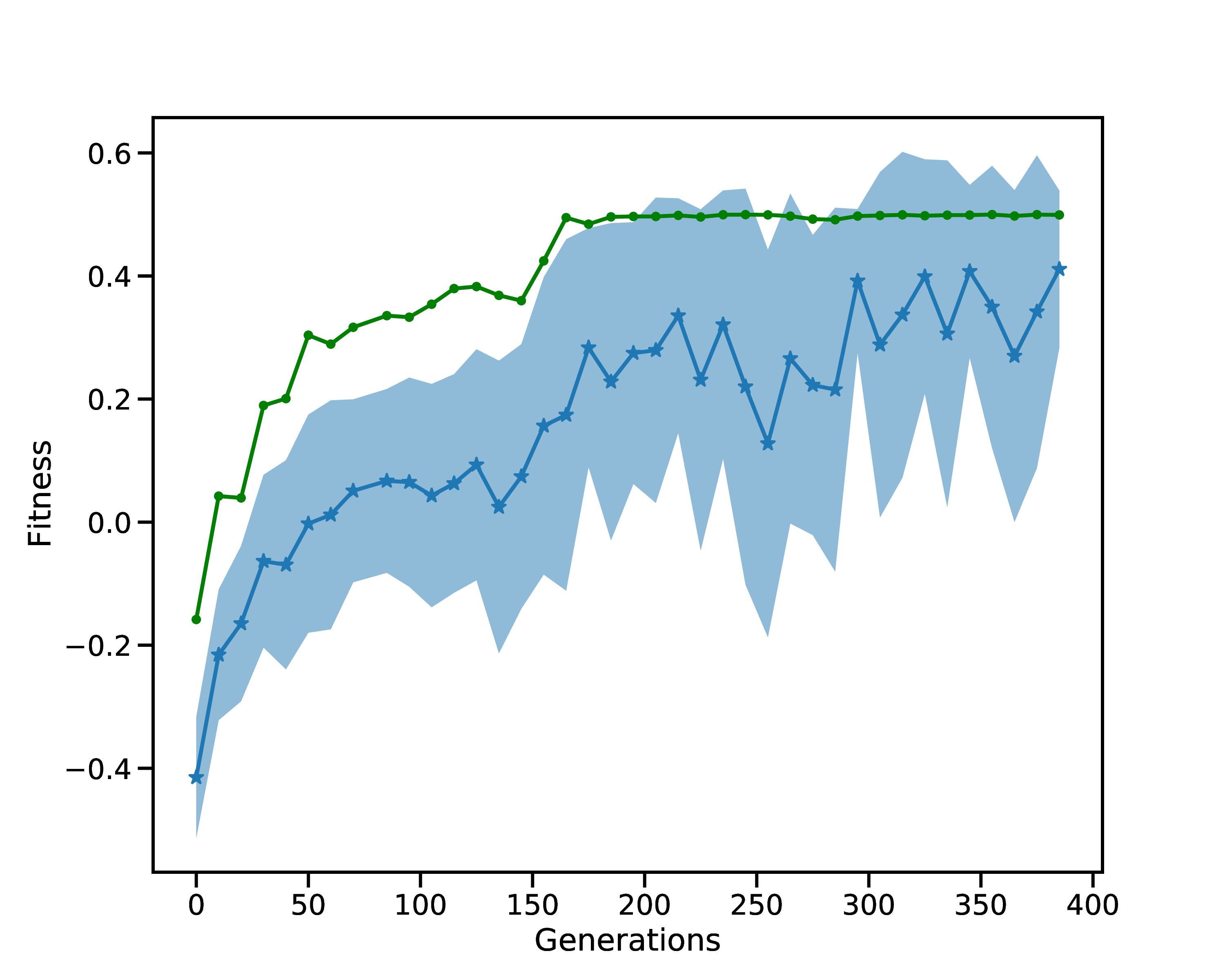}
  \caption{\label{fig:fitness_mc_fitness} Fitness of the MC run. The blue, starred line shows the mean of all individuals while the shaded area is the standard deviation. The green line is the best fitness per generation.}
\end{figure}
\subsubsection{Optimizer: Genetic algorithm}\label{subsubsec:uc5_outer}

The optimization method is identical to that used in Sec.~\ref{subsec:uc3_netlogo}.
Afterwards, the optimized parameters are sent back to the optimizee, which then initializes the next generation of individuals.

\subsubsection{Analysis}\label{subsubsec:uc5_results}

Figure~\ref{fig:fitness_mc_fitness} depicts the fitness of the SNN over $400$ generations.
After $50$ generations the fitness becomes positive showing that the car is moving towards the goal position. The best solution (goal position of $0.5$) is first reached around generation $160$.
In following generations the mean fitness saturates at around $0.3$, while the best fitness reaches the maximum of $0.5$.
After $400$ generations the L2L run is stopped with the best individual fitness being $0.5$.
Finally, we confirmed that the fittest individual could solve the MC problem. We ran a thousand episodes (each episode lasting for a maximum of $200$ simulation steps); the spiking network achieved the required average of 101 simulation steps to reach the goal position and thus solves the task (data not shown).

The MC problem has been approached using several machine learning techniques most of them focusing on reinforcement learning~\citep{clustered_rl2021, noisymc2008} and gradient descent~\citep{Metatrace2019} algorithms. Current implementations are able to solve the challenge while delivering a good performance in terms of speed of convergence and the obtained final score. We took an evolutionary approach by using a GA to optimize an SNN that is able to solve the MC obtaining consistently a high reward (over 100 trials). Evolutionary strategies have shown comparable performance to reinforcement learning and gradient descent algorithms in problems where learning to sense and act in response to the environment is required~\citep{design_nn_neuroevo, salimans2017evolution, Such2017DeepNG}. Another advantage with the evolutionary approach is the parallel exploration of the solution space. In L2L each individual is run as an independent optimizee process. The framework enables us to execute a large number of parallel optimizees in multi-core CPUs and HPC-infrastructures.

\subsection{Optimizing structural plasticity in NEST}\label{subsec:uc6_plasticity}
Structural plasticity is the ability of neurons to change their morphology in order to adapt to stimuli and reach specific activity regimes.
It depends on the ability of each neuron to create and remove synaptic connections with other neurons in the network.
It is a fundamental component of brain development, learning, healing after lesions and adaptation.
A network model with structural plasticity will change its connectivity through time, depending on a set of rules which are local to each neuron and regulate the generation or deletion of abstracted boutons and spines.

Note that the structural plasticity algorithm searches the connectivity space of the whole network in a multi objective optimization process, where all neurons aim at reaching a specific firing rate by creating an appropriate set of connections.
Structural plasticity works as a controller of the structure of the network aiming at reducing the error between the target activity set point of each neuron and its current activity.
The definition of these control guidelines is consolidated in the growth rules, thus their shape is very important to achieve the desired target.
The growth curve determines how fast and how strongly the structural changes will take place.
An inappropriately growth curve can lead to high oscillations in the activity of the network, taking it to an unstable regime.
Delayed structural changes with respect to the error signal can also result in a system that never converges to stable activity.

The goal of this task is to use L2L to optimize the growth rules that guide the structural plasticity algorithm implemented in NEST, which is based on a homeostatic principle: neurons that have an average activity below their set point will create new connections, and over-active neurons will delete them, according to a user-defined growth curve.
Frequently used growth curves are of linear and Gaussian shape.
Here, we make use of a Gaussian growth curve, which are defined by three parameters: the target firing rate $\epsilon$, the minimum start firing rate $\eta$, and the growth rate $\nu$.
The variable to optimize for each growth curve in this case is the growth rate $\nu$.

\subsubsection{Description of the simulation tools}\label{subsubsec:uc6_description}
In this use case we use NEST (see Section 3.1.1 of the main manuscript) and its implementation of structural plasticity.
This implementation is based on the model by \citet{butz2013simple} in which neurons are able to grow and rescind synaptic elements representing boutons and spines.
Compatible synaptic elements can be chosen to create a new synapse when they become mature.
If a synaptic element in an existing synapse is rescinded, then the synapse is deleted but the counterpart element can be rewired in a new synapse.
The growth of the synaptic elements is guided by homeostatic rules which can be defined by the user and can differ between populations or even neurons.
A comprehensive description of the structural implementation algorithm can be found in~\citet{diaz2016automatic}.

\subsubsection{Optimizee: Spiking neural networks with structural plasticity }\label{subsubsec:uc6_inner}

\begin{figure}
  \begin{center}
    \includegraphics[width=0.6\linewidth]{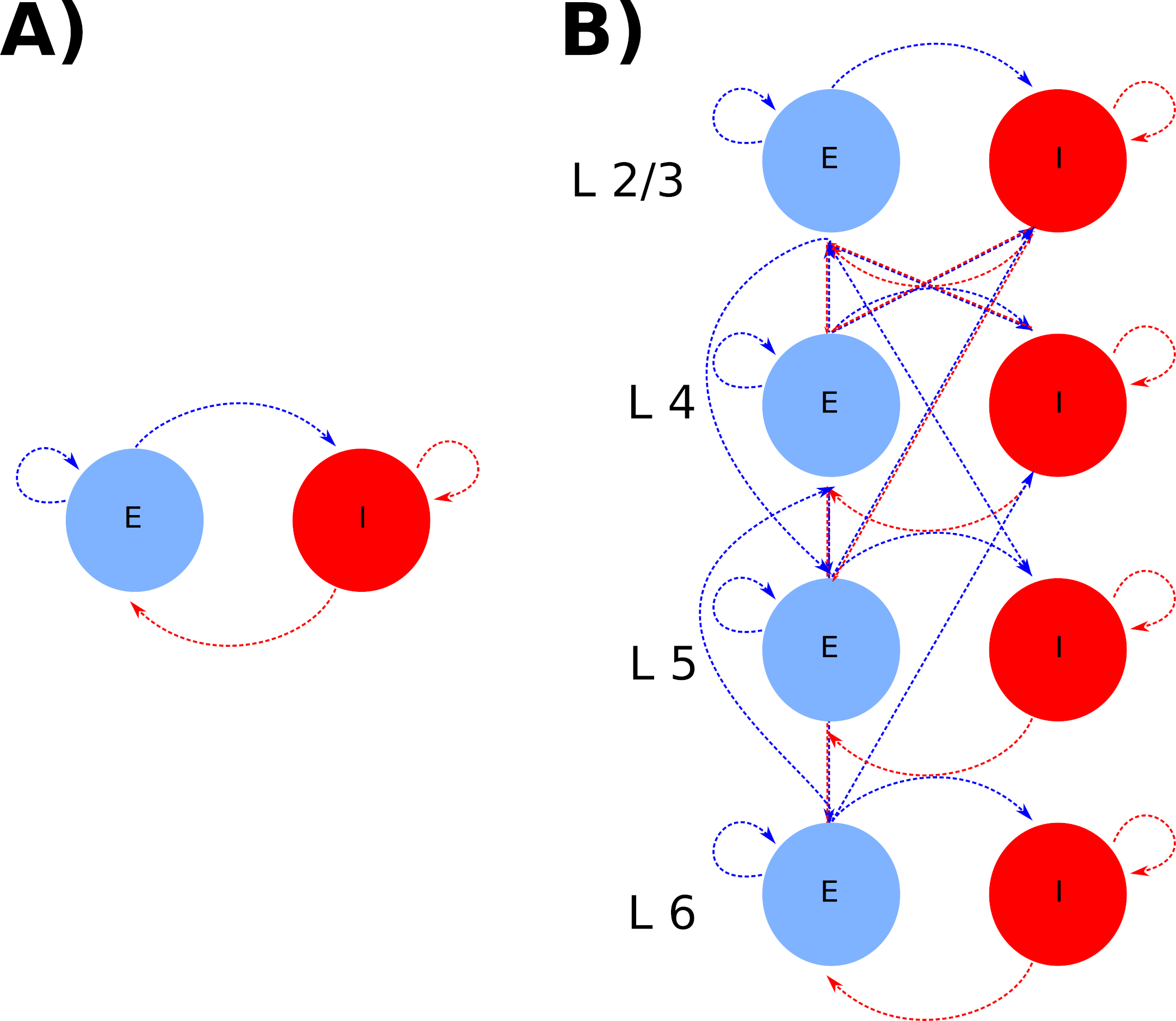}
  \end{center}
  \caption{
  Spiking neural network models with structural plasticity. \textbf{A)} Simple two population model, and \textbf{B)} the cortical microcircuit model.}\label{fig:models}
\end{figure}

In this use case we have two structural plasticity optimizees. The first optimizee consists of a simple two population model: the inhibitory population contains $20\%$ of the neurons in the network while the excitatory population contains the other $80\%$ (see Figure~\ref{fig:models}A.
The second optimizee is an implementation of the cortical microcircuit model proposed by \citet{potjans:2012}.
This model has eight populations, four excitatory and four inhibitory, in a layered fashion representing the cortical layers $2/3$, $4$, $5$, and $6$ (see Figure~\ref{fig:models}B)).

In the case of the two population model, the network is simulated for a total of $200\;\mathrm{cycles}$ of $2\;\mathrm{s}$ of biological time each. The task is to find a value for the growth rate $\nu$ which enables the network to reach target activity ranges for both populations within that time: between $5$ and $15\;\mathrm{spks/s}$ for the inhibitory population and between $1$ and $4\;\mathrm{spks/s}$ for the excitatory population.

For the microcircuit model optimizee the network is simulated for a total of $500\;\mathrm{cycles}$ of $1\;\mathrm{s}$ of biological time each. The task is to find growth rates for excitatory and inhibitory neurons ($\nu_\mathrm{e}$ and $\nu_\mathrm{i}$) which allow all populations to grow connections bringing them to their desired activity regimes within this time.
Each population has a different target firing rate according to the experimental recordings used to constrain the original model, see \citet{potjans:2012}.

Each job run provides a parallel set of initial states. Inner loop simulations are $200\;\mathrm{s}$ long because the network starts without any connections besides a Poissonian background noise and the timescale for structural plasticity is slow (order of seconds in our implementation) with respect to other neuronal timescales such as the membrane time constant.
At the end of each simulation, the average firing rate of all neurons is collected in order to evaluate the fitness, see below.

All experiments were performed using NEST~2.18 \cite{nest2_18} on the JURECA supercomputer of the Jülich Supercomputing Centre in Jülich, Germany.
\\

\subsubsection{Fitness metric}\label{subsubsec:uc6_metric}
In order to assess the performance of each instance of the network with specific parameters, a multiobjective fitness rule was derived.
Several compound fitness measures were tested for the multi-objective optimization, resulting in a weighted measure which gives more importance to the fitness of the excitatory than the inhibitory neurons because of their larger presence in the network.

The fitness metric used for the simple model is defined as:
\begin{equation}
  f = \frac{1}{0.8*(\lambda_\mathrm{e} - \epsilon_\mathrm{e}) + 0.2*(\lambda_\mathrm{i} - \epsilon_\mathrm{i})} \, ,
  \label{eq:sp1}
\end{equation}
where $\lambda_{\mathrm{e/i}}$ is the average firing rate of the excitatory (inhibitory) population at the end of the simulation and $\epsilon_{\mathrm{e/i}}$ is the corresponding target firing rate.
The choice of growth rate $\nu$ influences the number of connections created and the general activity of the network, leading to different simulation times.
In order to constrain the simulations, the maximum time allowed was set to $20\min$ and simulations which did not finish (i.e. successfully simulate $2\;\mathrm{s}$ of biological time) within that period were penalized with the worst fitness (a value of 0).

In the case of the microcircuit optimizee, a compound rule similar to the one in the simple model was used:
\begin{equation}
  f = \frac{1}{ \sum_{l} (\lambda^l_{\mathrm{e}} - \epsilon^l_{\mathrm{e}}) + (\lambda^l_{\mathrm{i}} - \epsilon^l_{\mathrm{i}}) } \, ,
  \label{eq:sp2}
\end{equation}
where $\lambda^l_{\mathrm{e/i}}$ and $\epsilon^l_{\mathrm{e/i}}$ are the average (of all neurons in a population at the end of the simulation) and target firing rates, respectively in the excitatory (inhibitory) population of layer $l$, and $l \in \{2/3,4,5,6\}$. As with the two-population model, a time limit was placed on the simulations. Any simulation that did not successfully simulate X biological seconds in $60\min$ of simulation was evaluated to have a fitness of $0$

\subsubsection{Optimizers}\label{subsubsec:uc6_outer}
For the simple two populations optimizee, the outer loop optimization algorithms were evolutionary strategies, gradient descent and simulated annealing.
For the microcircuit model optimizee, the outer loop algorithm chosen was gradient descent.

\subsubsection{Analysis}\label{subsubsec:uc6_results}
\begin{figure}[!ht]
  \begin{center}
    \includegraphics[width=\linewidth]{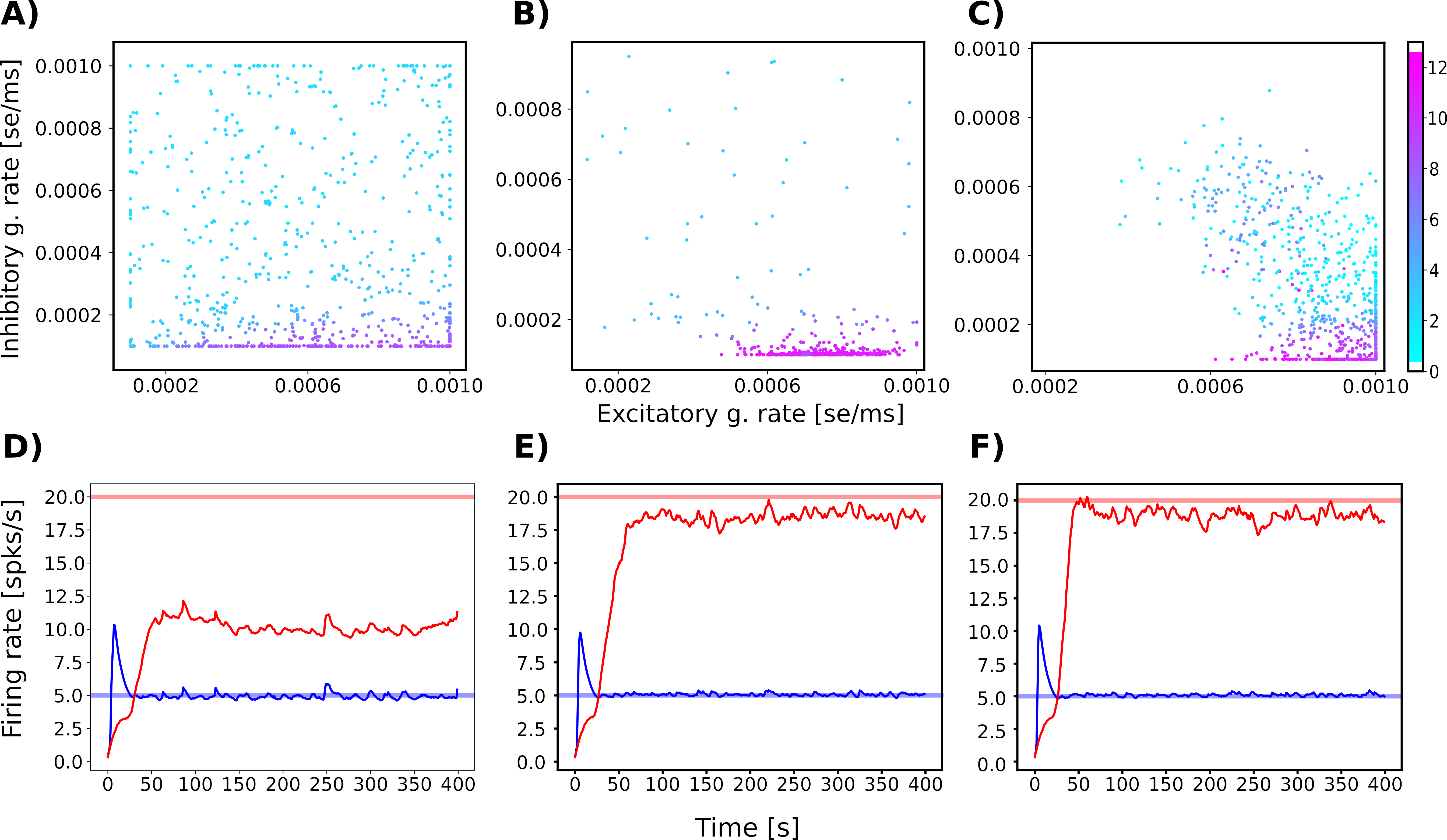}
  \end{center}
  \caption{
  Parameter exploration for the two population optimizee using \textbf{A)} gradient descent, \textbf{B)} simulated annealing and \textbf{C)} cross entropy as outer later optimizers.
  Each dot in the plot represents a position in the parameter space described by the inhibitory growth rate on the y-axis and the excitatory growth rate in the x-axis.
  The color of the dot indicates the fitness of that parameter combination.
  The performance of the best individual using gradient descent at generations $1$, $5$ and $20$ can be seen in panels \textbf{C)}, \textbf{D)}, and \textbf{E)} correspondingly.
  Blue and red curves represent the average firing rate of the neurons in the excitatory and inhibitory populations accordingly.
  The light blue and red horizontal lines represent the  target firing rates for the excitatory and inhibitory populations, respectively.
  }
\label{fig:generations}
\end{figure}
Figure \ref{fig:generations} shows the results of executing the simple optimizee in different scenarios.
In panel A) we can see that the simulated annealing algorithm performs a homogeneously distributed search on the space.
This has the advantage that local minima are easier to avoid.
The algorithm still focuses its search efforts in the lower right corner of the parameter space but uses some of the exploratory resources on areas of lower interest.
In the case of structural plasticity, simulated annealing is a good candidate to perform initial explorations of vast parameter spaces which provide a good idea of the shape of the space and where areas of interest are located.
The cross entropy algorithm is very effective in finding the area of interest very fast as shown in panel B).
This makes best use of computational resources but might miss other interesting areas of the parameter space to be explored.
Finally, the gradient descent algorithm, as shown in panel C) presents a middle ground between simulated annealing and cross entropy in terms of exploration and focus.
In all three algorithms we can see that there are parameter combinations outside of the lower right corner which have high fitness.
The random nature of the structural plasticity algorithm and the task definition can also cause variability in the performance of the model and thus make the exploration more challenging.
The figures~\ref{fig:generations}D), E) and F) show the best individual for generations $1$, $5$, and $15$ using gradient descent. The network converges to better solutions more quickly as the gradient descent algorithm finds more effective structural plasticity growth parameters.

\begin{figure}[!ht]
  \begin{center}
    \includegraphics[scale=0.7]{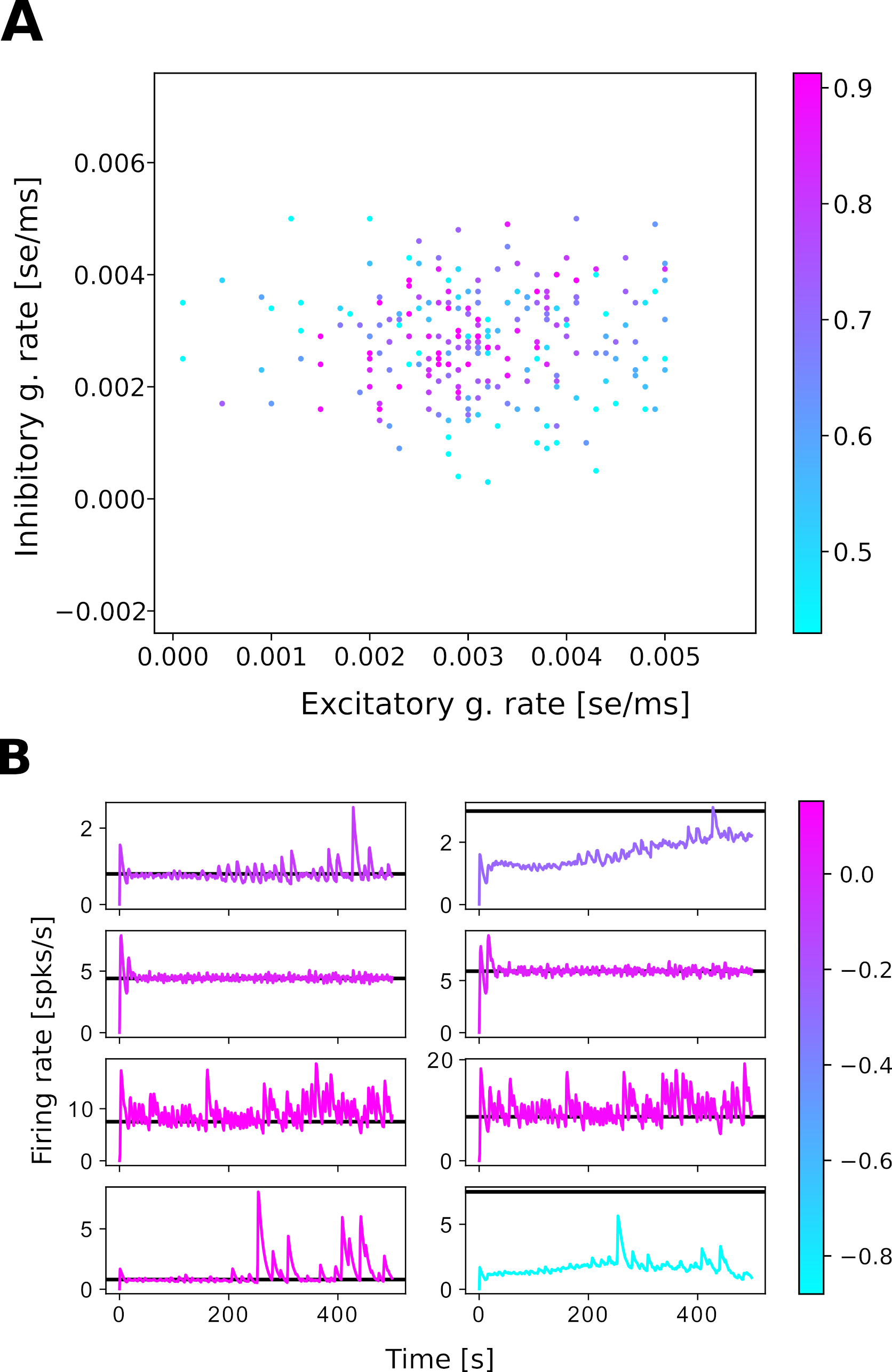}
  \end{center}
  \caption{A) Parameter exploration for the cortical microcircuit task.
  B) Best individual during the parameter exploration for the cortical microcircuit task after $25$ generations. The position of each subpanel corresponds to the position of each population in the model diagram in Figure~\ref{fig:models} B. The color of the line reflects the normalized difference between the average firing rate in each population at the end of the simulation and the target firing rate. The black horizontal lines mark the target firing rate of each population. }\label{fig:micro}
\end{figure}

In Figure~\ref{fig:micro} A) we can see the parameter exploration for the microcircuit model. Panel B) shows the best individual for the parameter exploration for the microcircuit model at generation $25$.
The best individual in generation $25$ has an inhibitory growth rate of $0.0029\:\mathrm{ se/ms}$ $\mathrm{(synaptic\ elements/ms)}$ and an excitatory growth rate of $0.0030\:\mathrm{ se/ms}$.
The results show that this model benefits from having growth rate values between excitatory and inhibitory neurons which are more similar to each other.
Gradient descent is an algorithm which is able to explore well the parameter space of the growth rules but also enables fast convergence.
The parameter space in this case seems to be smooth because the fitness increases gradually towards the point of best performance without many irregularities.
Because of this, gradient descent proves to be a good choice.

The simple metrics used in these use cases worked well but more complicated models might need to take into account other features of the network activity and normalize with respect to the target rate of each population.

The best parameter configuration depends on the flexibility we give to the model (e.g.~which connections are plastic and which network / cell parameters can be changed) as well as the functional and structural constraints we provide to the outer loop algorithm (e.g.~limits on total numbers of synapses per neuron or minimum ration between excitation and inhibition).
This is a more complex model which could benefit from exploring some variations.
For example, we could use the power spectrum of the activity in the model to define the fitness.
Different plasticity rules per layer could also be defined to enhance the flexibility of the model.
The L2L framework provides a platform to do large parameter explorations with adaptive algorithms, reducing time compared to brute force approaches.
Visual exploration of plasticity parameter spaces, as discussed in~\citet{nowke2018toward}, is very useful to find general areas of interest and understand the relationships between variables.
However, once this understanding is acquired, it can be further translated into good fitness functions which allow for a systematic exploration of the space with high throughput on HPC.
Structural plasticity simulations require long simulation times because the structural changes take place at a slow rate.
This is particularly critical as the networks become more complex or other plasticity rules are also in place.
Finding optimal ways to explore the parameter space is crucial to reduce wasting computational resources on areas of no interest or biological relevance.

\section{Discussion and Future Work}\label{sec:discussion}
Simulations in different science domains tend to become more and more complex and span over multiple disciplines and scales.
These simulations usually have a large number of parameters to configure, and researchers spend a long time tuning the model parameters manually, which is difficult and time-consuming.
To tackle these issues, it is necessary to have an automated tool which can be easily executed on local machines or likewise on super-computers.
We present the Learning to Learn framework as a flexible tool to optimize and explore ranges of parameter spaces.
Because the tool does not require a particular type of simulation, i.e.~it is agnostic to the model in the inner-loop, it enables the optimization of any type of parameter resulting from the model, as long as a fitness can be calculated and sent to the outer loop.

In Section~\ref{sec:results} we described several neuroscientific use cases at different scales.
The optimizations range from finding the correct set of parameter configurations, to determining network dynamics to solving optimization problems, up to exploring values for specific growth rules.
In all cases the optimization methods in the outer loop treated the inner loop simulations as black box problems and similarly the optimization technique was unknown to the inner loop.

In terms of implementation, every optimizee follows the same structure by providing three functions:
1.~creating the individual, i.e.~the parameters to optimized, 2.~starting and managing the optimizee run as well as providing a fitness to asses the simulation performance and 3.~optionally constraining the parameter exploration range.
The framework offers a plethora of built-in optimization techniques.
Most of them are population based optimizers, which require several individuals and a fitness or a fitness vector.
Both the fitness as well as the population approach are incorporated into the optimization.
For example, with genetic algorithms and the ensemble Kalman filter, the fitness is used to rank the individuals.
A large population enables a wider range to explore parameters and find possible good initializations, which leads to a faster convergence.
In order to not to get stuck in local optima, most of the optimizers offer techniques to perturb the individuals and additionally enlarge the parameter space (which of course can be bounded if needed).

Clearly, executing a high number of individuals leads to an increase in computational requirements.
By utilizing MPI in combination with the JUBE back-end, it is easy to deploy simulation and optimization on high performance computers in an automated fashion.
From the users perspective, only a few parameters have to be configured in a run script.
The optimizees for the inner loop are created and the simulations are executed in parallel.
One of the practical reasons of the population based optimizers is that the simulations are very easily parallelizable: each simulation can be conducted independently.
Only the parameters have to be collected in a single step and fed into the optimizer.
Afterwards, the optimized parameters are distributed for the next generation and the new simulations can be started.

The TVB use case is an example of demonstrating a parallelized simulation in the optimizee.
We show that we successfully reconfigured the gradient ascent optimizer to a version which can process a vector of fitnesses.
We used this optimizer to find the best parameter setting for a TVB model such that the match between simulated functional- and structural connectivity is optimal.
Results from performance testing for the RateML~\citep{rml} models show that for a double state model such as the \texttt{G2DO}, on a GPU with $40$ GB of memory, up to $\approx 62,464$ ($61$ times more parameters), can be simulated in a single generation, taking approximately the same amount of wall time due to the architecture of the GPU.
This would reduce the time it takes for each generation and increases the range and resolution of the to be optimized process even further; opening up possibilities for experiments requiring greater computational power.
Moreover, this particular optimizer is not limited to TVB simulations only.
Any process which uses a parallel architecture, e.g. GPU, CPU or FPGA, for which the output is a vector of fitnesses, can be adapted as an optimizee for the MGA optimizer.
The utilization of the \texttt{subprocess} library and information transfers via in- and output text files, makes usage of this optimizer generic for any process.
The MGA is just one example of an optimizer adapted to process multiple fitnesses, in theory any of the optimizers can be adjusted to handle multi fitness optimizees.

One important point to mention is the challenge of creating the fitness function.
Every fitness function is problem specific, and finding a suitable function is often a complex task.
In some cases the fitness is given by the design of the problem (c.f.~Section~\ref{subsec:uc1_nest}, in this case supervised learning).
To illustrate the point, the task in Section~\ref{subsec:uc3_netlogo} can be extended so that the ants are punished whenever they collide.
However, just adding a simple cost value for the collision makes the training and optimization much harder, the ants exhibit erratic behaviors such as spinning around or stopping moving after a few steps. Potentially, this behavior might resolve with enough generations, but it is more likely that the fitness function would need to be adapted. Even for the simple example shown here,  the fitness function had to be carefully balanced in terms of the punishment and reward cost, which lead to several trials and manual adjustments.
Thus, the explorative and exploitative behavior is influenced by the fitness function.
With a strict fitness function, i.e. every action in the simulation generates a reward or a punishment, it may be possible to exploit local optima, however it may restrict exploration of different, better optima.
Vice versa, making the fitness function too lax may lead to an overly exploratory behavior that doesn't exhibit any exploitation.

Specifically regarding our presented use cases, future work will include multi-objective optimization to decouple the objectives from a specific fitness function and optimize the fitness functions in interchangeable steps.
The L2L framework already supports multi-objective optimization since it can handle several fitness values.
Alternatively, the optimizee can be written in such a way that it exchanges the fitness function in certain generations and still returns one fitness value.

A visualization of the trajectories through generations may give further insights for a follow-up analysis of the parameters.
We aim to implement a visualization tool which can plot the evolution of the parameters using simple diagrams such as histograms, correlations and similar statistics. A desirable feature would be to interact with the plot while the simulation is ongoing, as demonstrated by Tensorboard\footnote{\url{https://www.tensorflow.org/tensorboard/}}.
A challenge here is to interact with the results whenever the run is conducted on a HPC, as\@
many super computing centers no longer allow X-forwarding - a network protocol to control and display a remote software from a local computer.
Instead, other mechanisms for interactive computing need to be considered such as Virtual Network Computing~\footnote{\url{https://trac.version.fz-juelich.de/vis/wiki/vnc3d}}.

In preliminary work, we already were able to run the L2L simulations on a HPC while instructing the run from a local machine.
By utilizing UNICORE~\citep{streit2005unicore}, a tool for distributed computing, we could successfully send an optimizee to a specified HPC, initialize the L2L framework, run the optimizations and collect the results.
For this approach to work, we have to ensure that the L2L framework is correctly deployed on the remote side.
A seamless integration of all tools in the process chain is required.
This approach also leads towards a vision of L2L as a service, where users can submit optimization workloads using a simple API.
Despite the advantages of this approach, new aspects should be considered to protect user data and any sensitive  data that can be used or produced during simulations.
In order to deploy this service, a full integration with the EBRAINS~\footnote{\url{https://ebrains.eu/}} infrastructure is our target for the near future, as this will enable L2L to support the neuroscience community while being part of a well-established research platform.

Another necessary element, which currently only available in a preliminary form, is check-pointing the run, i.e.~the possibility to continue the inner and outer loop process to a later time. This would allow us to execute jobs in a very long period without any HPC time restriction.
At the moment the run-script (see Section~\ref{subsec:l2l_workflow}) has to be changed with a few more routines to load the trajectories from an earlier run and to continue it.
In an upcoming release this component will be integrated into the L2L framework.

Finally, we would like to extend the set of optimization techniques with optimizers which have more capabilities.
This would be for example a neural network, along the lines of the approach proposed by~\cite{andrychowicz2016learning}.
For instance, the network could learn the distribution of the parameter space and predict the next set of parameters.
One other interesting direction is to include Bayesian Optimization via Bayesian hierarchical modelling.
In this case the parameters are not optimized directly as depicted in this work, instead uncertainty measures and prediction uncertainty are inferred~\citep{gordon2018meta,finn2018probabilistic, yoon2018bayesian}.

In conclusion, with this work we have presented L2L as a software framework for the hyper-parameter optimization of computing workloads, specially focusing on neuroscience use cases. The flexibility of this framework is designed to support the broad and interdisciplinary nature of brain research and provides easier access to high performance computing for machine learning based optimization tasks.

\bibliographystyle{unsrtnat}
\bibliography{l2l.bib}  






\end{document}